%% file: main.tex
\documentclass{article}
\usepackage[accepted]{aistats2025}

\usepackage{comment}
\usepackage{graphicx}

\input{math_commands.tex}

\usepackage{hyperref}
\usepackage{url}

\usepackage[round]{natbib}
\renewcommand{\bibname}{References}
\renewcommand{\bibsection}{\subsubsection*{\bibname}}

\newcommand{\papertitle}{}
\renewcommand{\papertitle}{Domain Adaptation and Entanglement: an Optimal Transport Perspective}
\newcommand{\shorttitle}{}
\renewcommand{\shorttitle}{Domain Adaptation and Entanglement}

\newcommand{\new}[1]{\marginpar{\color{blue} NEW}{\color{blue} #1}}

\newcommand{\fix}[1]{\marginpar{\color{red} FIX}{\color{red} #1}}

\newcommand{\removepf}[1]{#1}

\renewcommand{\#}{\sharp}

\begin{document}

\onecolumn

\runningtitle{\shorttitle}

\aistatstitle{\papertitle}

\aistatsauthor{Okan Ko\c c \And Alexander Soen}
\aistatsaddress{RIKEN AIP \And RIKEN AIP \& Australian National University}
\aistatsauthor{Chao-Kai Chiang \And Masashi Sugiyama}
\aistatsaddress{The University of Tokyo \And The University of Tokyo \& RIKEN AIP}
\runningauthor{Okan Ko\c c, Alexander Soen, Chao-Kai Chiang, Masashi Sugiyama}

\input{sections/abstract}

\input{sections/introduction}

\input{sections/related}
\input{sections/notation}
\input{sections/base_theory}
\input{sections/entanglement}
\input{sections/analysis}

\input{sections/experiments}
\input{sections/conclusion.tex}

\acknowledgments{
Chao-Kai Chiang was supported by the Institute for AI and Beyond, U. Tokyo.
Masashi Sugiyama was supported by the Institute for AI and Beyond, U. Tokyo and by a grant from Apple, Inc. 
Any views, opinions, findings, and conclusions or recommendations expressed in this material are those of the authors and should not be interpreted as reflecting the views, policies or position, either expressed or implied, of Apple Inc.
}

\bibliographystyle{apalike}
\bibliography{paper.bib}

\input{appendix}

\end{document}

%% file: math_commands.tex
\usepackage[latin1]{inputenc}
\usepackage{amssymb}
\usepackage{amsthm}
\usepackage{amsmath,amsfonts,bm}
\usepackage{xcolor}
\usepackage{stmaryrd}
\usepackage{booktabs,multirow}

\newtheorem{theorem}{Theorem}[section]
\newtheorem{corollary}{Corollary}[section]
\newtheorem{lemma}{Lemma}[section]

\theoremstyle{definition}
\newtheorem{definition}{Definition}[section]

\newtheorem{assumption}{Assumption}[section]

\newcommand{\tr}{\mathrm{tr}}

\newcommand{\risk}[0]{\epsilon}
\newcommand{\expect}[0]{\mathop{\mathbb{E}}}

\newcommand{\dd}{\mathrm{d}}

\newcommand{\KL}{\mathrm{KL}}
\newcommand{\JS}{\mathrm{JS}}

\newcommand{\LJE}{\mathrm{LJE}}
\newcommand{\CCA}{\mathrm{CC}}
\newcommand{\GS}{\mathrm{GS}}

\newcommand{\entangle}{\mathcal{E}}
\newcommand{\labelentangle}{\entangle_{\rm y}}
\newcommand{\predentangle}{\entangle_{\rm \hat{y}}}

\newcommand{\Mid}{\; \Vert \;}

\newcommand{\oeub}{\mathcal{U}}
\newcommand{\wrr}{\mathcal{R}}

\newcommand{\defeq}{\doteq}

\usepackage{xspace}
\makeatletter
\DeclareRobustCommand\onedot{\futurelet\@let@token\bmv@onedotaux}
\def\bmv@onedotaux{\ifx\@let@token.\else.\null\fi\xspace}
\def\eg{e.g\onedot} 
\def\ie{i.e\onedot}

\def\wrt{w.r.t\onedot}

\def\bigoh{\mathcal{O}}
\makeatother

\newcommand{\snegativespace}{\vspace{-0.15cm}}

%% file: sections/abstract.tex
\begin{abstract}
    Current machine learning systems are brittle in the face of distribution shifts (DS), where the target distribution that the system is tested on differs from the source distribution used to train the system. This problem of robustness to DS has been studied extensively in the field of domain adaptation. For deep neural networks, a popular framework for unsupervised domain adaptation (UDA) is domain matching, in which algorithms
    try to align the marginal distributions in the feature or output space.
    The current theoretical understanding of these methods, however, is limited and existing theoretical results are not precise enough to characterize their performance in practice. 
    In this paper, we derive new bounds based on optimal transport that analyze the UDA problem. Our new bounds include a term which we dub as \emph{entanglement}, consisting of an expectation of Wasserstein distance between conditionals with respect to changing data distributions. Analysis of the entanglement term provides a novel perspective on the unoptimizable aspects of UDA. In various experiments with multiple models across several DS scenarios, we show that this term can be used to explain the varying performance of UDA algorithms.
    %
\end{abstract}

%% file: sections/introduction.tex
\section{Introduction}

Deep learning has revolutionized machine learning and AI in the past decade and provided impressive solutions in computer vision, speech recognition and natural language processing, among others~\citep{bengio2017deep}. 

However one of the reasons preventing their wider deployment is the trustworthiness and robustness issues surrounding deep neural networks (DNN). DNNs are often not robust to distribution shifts that occur in many critical environments, such as those encountered in robotics, healthcare and computer vision, to name a few.

The generalization guarantees of standard supervised learning, typically used to train DNNs, require the training and test data to come from the same distribution~\citep{shalev2014understanding}.
%
The performance of DNNs is often brittle whenever this assumption is violated, however, and degrades quickly in the face of even mild distribution shifts at times. Domain adaptation studies this robustness to distribution shift, in cases where the target distribution, from which unlabeled samples are available, is different from the source distribution that the model is trained on.

In this paper, we are interested in target risk upper bounds for \emph{unsupervised} domain adaptation (UDA), in which there is no labeled data available from the target domain. In particular, we introduce novel upper bounds based on optimal transport~\citep{villani2009optimal, peyre2019computational}, and study them from the
optimization perspective. Based on the unoptimizable part of the upper bound (which requires target labels), we introduce the notion of \emph{entanglement}\footnote{Not to be confused with the notion of disentangled representations \citep{bengio2013representation} in representation learning research.} which quantifies the central difficulty of UDA from the optimization perspective. Unlike previous bounds~\citep{benDavid2010, courty2017}, the entanglement term makes explicit and quantifies the \emph{hardness} of UDA: algorithms that minimize marginal distance (or divergence) can entangle different class-conditionals while pulling unlabeled samples closer together, making it hard (and often impossible) to find a good classifier that has high target accuracy. Previous work using the accuracy of the best model in source and target domains to bound the target accuracy hides this difficulty because this quantity can grow as the model is optimized~\citep{johansson2019support}. Experiments confirm that the entanglement term is a suitable tool to analyze the UDA performance of many algorithms: depending on the (model, dataset, optimizer) triplet this term can grow during the optimization and cause target accuracy to degrade. Throughout the paper, by making various assumptions, such as on the loss function or the model capacity, we study the entanglement term and the resulting UDA problem in detail. Extensive experiments with different models, datasets and optimizers corroborate our conclusions.
%


\input{figures/teaser}

%% file: figures/teaser.tex
\begin{figure*}[t!]
    \centering
    \includegraphics[width=0.3\linewidth]{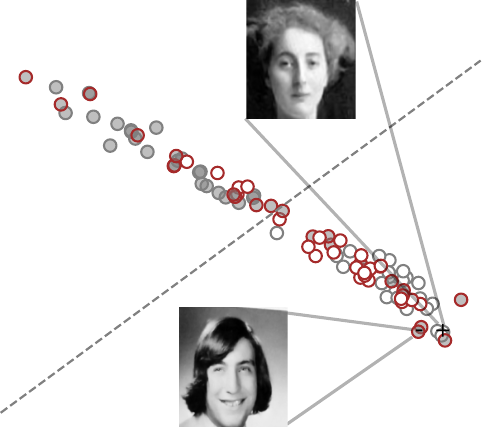}
    \caption{
    %
    %
    We derive new bounds for unsupervised domain adaptation based on optimal transport and introduce a term called entanglement, which quantifies the loss in accuracy that can happen when aligning marginals. 
    During the marginal alignment step, optimal transport associates source inputs (gray) to target inputs (brown) and stochastic gradient descent tries to find neural network parameters that pull these coupled points closer together. 
    Entanglement measures the average loss of associating pairs with \emph{different} labels.
    In this figure we visualize such an entangled (source, target) output pair using the Portraits dataset. Shaded circles correspond to images labeled female and empty circles to males. The dotted line indicates the decision boundary in the output space of a convolutional neural network separating male predictions from females.
    }

    \label{fig:portraits_entanglement}
\end{figure*}

%% file: sections/related.tex
\section{Related Work}\label{related}

Domain adaptation (DA) studies ways to transfer knowledge between similar \emph{domains}, which consist of a joint input-label space and a probability distribution over this space. To categorize very briefly the algorithms proposed in the literature (see, e.g., \citet{wilson2020survey} for a survey): the earliest algorithms considered using \emph{importance weighting} (IW) to adapt misspecified source models to the target distribution \citep{shimodaira2000improving, huang2006correcting}. Under the covariate shift and the equal support assumptions, one can show that importance weighting can be used to learn consistent models~\citep{sugiyama2012machine}. \emph{Domain matching} methods \citep{ganin2016, damodaran2018deepjdot, shen2018wasserstein, combes2020}, try to find representations that minimize the difference between the source and the target distributions, quantified usually by a divergence term between the two. \emph{Pseudolabeling}, widely used in the semi-supervised learning literature \citep{lee2013pseudo}, can also be used to perform domain adaptation~\citep{Shu2018ADA}, although without any guarantees, unless other assumptions are made \citep{kumar2020understanding}.

On the theoretical side, the seminal paper of \citet{benDavid2010} studies conditions under which a source-classifier does well on a related, but different target distribution. Under the condition that there exists a classifier that does well (i.e., has high accuracy) on both source and target domains, the knowledge transfer from source to target is then quantified by an $\mathcal{H} \Delta \mathcal{H}$ divergence that can be thought of informally as the projection of the total variation distance to the hypothesis space $\mathcal{H}$. Inspired by the theory, adversarial domain matching approaches~\citep{ganin2016} try to find representations that minimize this divergence together with the source risk minimization. Despite the popularity of adversarial losses, such approaches lack rigorous guarantees, due to the lack of invertibility~\citep{johansson2019support} of the models or more generally
the lack of invariant features~\citep{zhao2019learning, arjovsky2019invariant}.

A research direction in DA that is closely aligned with our work consists of changing the $\mathcal{H} \Delta \mathcal{H}$ divergence to various other dissimilarity functions. Some common dissimilarity functions considered are the KL-divergence~\citep{nguyen2022kl} and -- most relevant to this work -- Wasserstein distances~\citep{redko2017,courty2017}. The central property that these dissimilarity functions provide is the change of measure inequality, which allows the target risk to be upper bounded by a source risk plus the corresponding dissimilarity measure~\citep{donsker1983asymptotic}. These bounds inspired several algorithms to optimize deep neural networks~\citep{damodaran2018deepjdot, shen2018wasserstein}. We discuss the proposed algorithms and related work in more detail in Appendix~\ref{app:related_work}.



%% file: sections/notation.tex
\paragraph{Notation.} To set the notation for what is to follow: we are given input-label samples $(x,
y)$ from measurable metric spaces $(\mathcal{X}, d_x)$ and $(\mathcal{Y}, d_y)$.
For a sample space $\mathcal{Z}$, we denote the set of probability measures as $\triangle(\mathcal{Z})$.
We typically denote joint distributions as $p, q \in \triangle(\mathcal{X} \times \mathcal{Y})$. Their corresponding marginal distributions are denoted via a subscript, i.e., $p_{\rm x} \in \triangle(\mathcal{X})$ or $q_{\rm y} \in \triangle(\mathcal{Y})$.
With slight abuse of notation, we will overload $\mathcal{Y}$ for the corresponding simplex $\triangle(\mathcal{Y})$ --- which will become useful due to classifiers commonly outputting softmax probabilities.
%
We denote the pushforward measure of a measurable function $f \colon \mathcal{X} \rightarrow \mathcal{Y}$ and a probability measure $\mu \in \triangle(\mathcal{X})$ by $f\#\mu$.
With some abuse of notation, for a joint distribution $p \in \triangle(\mathcal{X} \times \mathcal{Y})$ we denote the pushforward $f \# p \defeq T_{(f, \mathrm{id})} \# p$, where $T_{(f, \mathrm{id})}$ is the map $(x, y) \mapsto (f(x), y)$.
%

%% file: sections/base_theory.tex
\section{Entanglement in Domain Matching}

In DA, we are interested in finding a model $f \colon \mathcal{X}
\rightarrow \mathcal{Y}$ from a hypothesis class $\mathcal{H}$ which transfers well to a target domain: the model should not only achieve high accuracy on the source $p \in \triangle(\mathcal{X} \times \mathcal{Y})$ but \emph{also} maintain similar levels of accuracy on a target distribution $q \in \triangle(\mathcal{X} \times \mathcal{Y})$.
We consider the setting where we only have access to full samples from the source $(x, y) \sim p$, but only have unlabeled data from the target, i.e., $y'$ is not observed in samples $(x', y') \sim q$. 
The performance on each of these distributions can be quantified via their corresponding risk. Given a loss function
$\ell \colon \mathcal{Y} \times \mathcal{Y} \rightarrow \mathbb{R}$, we define
the source and target risks, respectively, as
\begin{equation*}
    \risk_{p}(f) \doteq \hspace{-1.2mm} \expect_{(x, y) \sim p}[\ell(f(x),y)]; \;  
    \risk_{q}(f) \doteq \hspace{-1.2mm} \expect_{(x', y') \sim q}[\ell(f(x'),y')].
\end{equation*}
%
%
Motivated with the goal of keeping the target risk $\risk_{q}(f)$ small, in this
section we provide novel bounds that connect target risk minimization to source
risk minimization through optimal transport.

\paragraph{Optimal Transport.}
Amongst the set of distortion measures utilized in machine learning, optimal
transport distances~\citep{peyre2019computational,villani2009optimal} (using an appropriately chosen cost function) have been an appealing choice for comparing probability measures since (i) they satisfy the requirements of a metric between probability measures and (ii) they can be used between any two distributions $\mu, \nu$ regardless of their support (unlike $f$-divergences~\citep{ali1966general,csiszar1967information}).
\begin{definition}
    \label{def:p-wasser}
    Let $c \colon \mathcal{X} \times
    \mathcal{Y} \rightarrow \mathbb{R}$ be a (non-negative) cost function. The $\alpha$-Wasserstein distance \wrt $c$ between measures $\mu \in \triangle(\mathcal{X})$ and $\nu
    \in \triangle(\mathcal{Y})$ is given by,
    \begin{equation*}
        W_{\alpha, c}(\mu, \nu) \defeq \left( \inf_{\gamma \in \Gamma(\mu ,\nu)}\int_{\mathcal{X} \times \mathcal{Y}} c(x, y)^\alpha \dd\gamma(x, y) \right)^{1/\alpha},
    \end{equation*}
    where
    $
        \Gamma(\mu, \nu) 
        \hspace{-0.5mm} \defeq  \hspace{-0.5mm}
        \left\{ \gamma \in \triangle(\mathcal{X} \times \mathcal{Y}) 
        \hspace{-0.2mm}
        :
        \hspace{-0.2mm}
        \pi_{1}\#\gamma = \mu, \pi_{2}\#\gamma = \nu \right\},
    $
    and $\pi_{1}(x, y) = x$, $\pi_{2}(x, y) = y$ are projection functions for the corresponding marginal constraints.
\end{definition}

Typically, the cost function of a Wasserstein distance $c$ is implicit in the sample space of
$\mathcal{X}$ and $\mathcal{Y}$, i.e., when $\mathcal{X} = \mathcal{Y}$ is
a metric space, one uses the underlying metric $d_{x}$. In our work, we will make the choice of the cost function explicit. In particular, we are interested in Wasserstein distances over probability measures with \emph{joint sample spaces} when the cost function
\emph{decomposes} over the sample space's Cartesian product $\mathcal{X} \times \mathcal{Y}$.

\begin{definition}\label{def:decomp}
    We say that a cost function $c \colon (\mathcal{X} \times \mathcal{Y})
    \times (\mathcal{X} \times \mathcal{Y}) \rightarrow \mathbb{R}$ is
    decomposable if there exist cost functions $c_1 \colon \mathcal{X} \times
    \mathcal{X} \rightarrow \mathbb{R}$ and $c_2 \colon \mathcal{Y} \times
    \mathcal{Y} \rightarrow \mathbb{R}$ such that $c((x, y), (x', y')) = c_1(x,
    x') + c_2(y, y')$.
    We further denote the corresponding $\alpha$-Wasserstein distance for decomposable
    costs by $W_{\alpha, c_1, c_2} = W_{\alpha, c}$.
\end{definition}

It can be immediately seen that if a cost function decomposes into two metric cost functions $c_1, c_2$, then the original cost function must also be a metric cost function. This provides a simple method for defining a decomposable cost by the addition of metric functions $c_1 \colon (\mathcal{X} \times \mathcal{X}) \rightarrow \mathbb{R}$ and $c_2 \colon (\mathcal{Y} \times \mathcal{Y}) \rightarrow \mathbb{R}$.

\subsection{Loss Function and Model Assumptions}


%
In this section, we will establish bounds on the target risk $\risk_{q}(f)$ via Wasserstein distances of decomposable cost functions. In particular, the decomposable cost we will use will be a function of the loss function $\ell \colon \mathcal{Y} \times \mathcal{Y} \rightarrow \mathbb{R}$. Before establishing our bounds, we first state the assumptions needed on the loss function $\ell$ and classifier $f$.

\begin{assumption}[Metric Loss]
    \label{assum:metric}
    The loss function $\ell \colon \mathcal{Y} \times \mathcal{Y} \rightarrow \mathbb{R}$ is symmetric, non-negative (hence $\ell(x, y) \geq 0$ for all $x,y$), and respects the triangle inequality. In addition, the identity of indiscernibles, i.e., $\ell(y, y') = 0 \iff y = y'$ holds.
\end{assumption}

Assumption~\ref{assum:metric} is commonly made in the DA literature, in particular, the symmetry and triangle inequality properties of the loss function are often utilized to factor out terms involving the best model $f^{\star}$ (minimizing the source and target risks jointly)~\citep{benDavid2010, courty2017}. It should be noted that although symmetry can be accounted for via symmetrizing the loss function -- \eg, $\tilde{\ell}(y, y') = \frac{1}{2} (\ell(y, y') + \ell(y', y))$ -- not all symmetric loss functions will satisfy the triangle inequality. Despite this, Assumption~\ref{assum:metric} can be replaced with an \emph{approximate triangle inequality}~\citep{crammer2008learning} with slight alteration to our results. We discuss the extension of our results to non-metric loss functions in Appendix~\ref{app:approx_triangle_ineq} in detail. 

\begin{assumption}[Bounded Loss]
    \label{assum:bounded}
    The loss function $\ell \colon \mathcal{Y} \times \mathcal{Y} \rightarrow \mathbb{R}$ is bounded such that for all $y, y' \in \mathcal{Y}$, $\ell(y, y') \leq L$ for $L > 0$. 

    %
\end{assumption}

When $\mathcal{Y}$ consists of a finite set $\vert \mathcal{Y} \vert < \infty$ and $\ell$ satisfies the identity of indiscernibles,  then the loss function satisfies also a positive
\emph{lower} bound. Assumption~\ref{assum:bounded} and the lower bound correspond to Lipschitz conditions \wrt a discrete topology of $\mathcal{Y}$. We require the Lipschitz conditions only in Section~\ref{sec:entanglement} to analyze our bounds in the presence of label shift.



In addition to loss function assumptions, we also consider the following assumption for the models $f \in \mathcal{H}$.

\begin{assumption}[Surjective Models]
    \label{assum:surjective}
    The model $f \colon \mathcal{X} \rightarrow \mathcal{Y}$ is a surjective function.
\end{assumption}

The surjectivity simply states that model $f$ will map an input to all possible outputs. It can be relaxed to hold only for some non-empty subset of $\mathcal{Y}$, however the requirement should hold uniformly over the hypothesis class of $f$ whenever the bounds are minimized.

\subsection{Bounds on Target Risk}
We start with a simple, yet effective bound using suitable
Wasserstein distances that use a decomposable cost function based on the loss
function $\ell$. To simplify notation, we will denote $(\ell \circ f)(x, x') =
\ell(f(x), f(x'))$.
\begin{lemma}\label{lemma1}
    Suppose the loss function $\ell \colon \mathcal{Y} \times \mathcal{Y}
    \rightarrow \mathbb{R}$ satisfies Assumption~\ref{assum:metric}. Then the target \emph{risk} of a
    classifier $f \colon \mathcal{X} \rightarrow \mathcal{Y}$ is bounded by
    \begin{align}
        \risk_{q}(f) 
        &\leq \risk_{p}(f) + W_{1, \ell \circ f, \ell}(p, q) \overset{\rm (s)}{\leq} \risk_{p}(f) + W_{1, \ell, \ell}(f\#p, f\#q) \label{bound1},
    \end{align}
    where the inequality $\rm (s)$ holds for surjective $f$. Additionally, $\rm (s)$ is an \emph{equality} whenever $f$ is invertible.
    We remind the readers that 
    $(f\#p)(\hat{y}, y) = p( f^{-1}(\hat{y}), y)$.
\end{lemma}
\removepf{
\begin{proof}[Proof Sketch]
    The risk difference $\risk_{q}(f) - \risk_{p}(f)$ can be related to any coupling that integrates over the corresponding loss differences. The optimal transport cost is the lowest cost among such couplings and applying triangle inequality to its integrand yields the Wasserstein distance above. Surjectivity 
    allows us to relate the optimal transport between $p$ and $q$ to that between $f\# p$ and $f\# q$.
\end{proof}
}
It should be noted that the 1-Wasserstein distance in Lemma~\ref{lemma1} can be
easily replaced with a corresponding $\alpha$-Wasserstein distance for $1 \leq \alpha <
\infty$ due to the ordering of $\alpha$-norms, i.e., $W_{\alpha, c} \leq W_{\beta,
c}$ for any $\alpha < \beta$. This also indicates that out of the typical
ranges of $\alpha$-Wasserstein distances considered, the presentation of
Lemma~\ref{lemma1} taking $\alpha=1$ is the most tight.

The second term in \eqref{bound1} is the Wasserstein distance between the \emph{joint} distributions $p$ and $q$ when using the aggregate loss $\ell(f(x), f(x')) + \ell(y, y')$ between prediction-label pairs $(f(x), y)$ and $(f(x'), y')$. Since in UDA the labels $y'$ from the target distribution are unavailable, the bound above is not immediately useful in our case. Loosening the bound is inevitable in order to derive practical optimizers, hence we next relate the bound to the marginal Wasserstein distance $W_{1, \ell \circ f}(p_{\rm x}, q_{\rm x})$ which was used in several DA works~\citep{courty2017, shen2018wasserstein, redko2017}. 

In order to relate the \emph{joint} Wasserstein distance to Wasserstein
distances over marginal distributions, we consider the following lemma that
exploits the structure of decomposable joint costs.
\begin{lemma}\label{lemma2}
    Let $c \colon (\mathcal{X} \times \mathcal{Y}) \times (\mathcal{X} \times
    \mathcal{Y}) \rightarrow \mathbb{R}$ be a cost function that
    decomposes into functions $c_1 \colon \mathcal{X} \times \mathcal{X}
    \rightarrow \mathbb{R}$ and $c_2 \colon \mathcal{Y} \times \mathcal{Y}
    \rightarrow \mathbb{R}$.
    Then for $1 \leq \alpha < \infty$, 
    \begin{align}
        W_{\alpha, c_1}(\mu_{\rm x}, \nu_{\rm x}) \leq W_{\alpha, c}(\mu, \nu) \leq W_{\alpha, c_1}(\mu_{\rm x}, \nu_{\rm x}) +
        \expect_{(x, x') \sim \gamma_{x}^{\star}}\left[W_{\alpha, c_2}^{\alpha}(\mu_{\rm y \mid x}(\cdot \mid x), \nu_{\rm y \mid x}(\cdot \mid x')) \right]^{1/\alpha} \label{eq:lem-decomp-x}, \\
        W_{\alpha, c_2}(\mu_{\rm y}, \nu_{\rm y}) \leq W_{\alpha, c}(\mu, \nu) \leq W_{\alpha, c_2}(\mu_{\rm y}, \nu_{\rm y}) +
        \expect_{(y, y') \sim \gamma_{y}^{\star}}\left[W_{\alpha, c_1}^{\alpha}(\mu_{\rm x \mid y}(\cdot \mid y), \nu_{\rm x \mid y}(\cdot \mid y')) \right]^{1/\alpha}, \label{eq:lem-decomp-y}
    \end{align}
    where $\gamma_x^{\star}$ and $\gamma_y^{\star}$ are optimal couplings corresponding to $W_{\alpha,c_1}(\mu_x, \nu_x)$ and $W_{\alpha, c_2}(\mu_y, \nu_y)$, respectively.
%
\end{lemma}
\removepf{
\begin{proof}[Proof Sketch]
    To prove the lower bounds, we relate the optimal coupling between the joint distributions to the optimal coupling between the marginals by a transformation that disappears once we lower bound the joint cost $c$ with the marginal costs.
    The proof for the upper bounds hinges on the fact that transports that decompose the optimal coupling between $(x,y) \sim \mu$, $(x', y') \sim \nu$ pairs are sub-optimal: the couplings $\gamma(x,x') \gamma(y, y'| x, x')$ and $\gamma(y,y') \gamma(x, x'| y, y')$ incur a cost that upper bounds the optimal transport cost. 
\end{proof}
}
%

For the rest of the paper, we will need both directions of the bound\footnote{See Appendix \ref{add_remarks} where we explore the bounds in the case of Gaussians.} for further analysis. In particular, we will use the first bound \eqref{eq:lem-decomp-x} to analyze DA \emph{algorithmically}, while the second bound \eqref{eq:lem-decomp-y} will be a useful theoretical tool for that purpose. 

%% file: sections/entanglement.tex
\subsection{Entanglement}\label{intro_entanglement}

Next, we establish a corollary that links the two previous lemmas. The corollary introduces an \emph{entanglement} term between the source and target distributions, which fundamentally results from aligning marginal distributions during UDA without having access to target label information. First, we define two closely related entanglement terms.

\begin{definition}\label{def:entangle}
    For a model $f \colon \mathcal{X} \rightarrow \mathcal{Y}$, the label entanglement $\labelentangle(f)$ and the prediction entanglement $\predentangle(f)$ are defined as
    \begin{align*}
        \labelentangle(f) 
        &\defeq 
        \int_{\mathcal{Y} \times \mathcal{Y}} W_{1, \ell}(p_{\rm y \mid f}(\cdot \mid \hat{y}), q_{\rm y \mid f}(\cdot \mid \hat{y}')) \dd \gamma^\star_{\rm f}(\hat{y}, \hat{y}'),
        \\
        \predentangle(f) 
        &\defeq 
        \int_{\mathcal{Y} \times \mathcal{Y}} W_{1, \ell}(p_{\rm f \mid y}(\cdot \mid y), q_{\rm f \mid y}(\cdot \mid y')) \dd \gamma^\star_{\rm y}({y}, {y}'),
    \end{align*}
    where $p_{\rm y \mid f}(y \mid \hat{y})$ and $q_{\rm y \mid f}(y \mid \hat{y}')$ correspond to the joint distributions $p, q$  conditioned on $\hat{y} = f(x)$ and $\hat{y}' = f(x')$, while $p_{\rm f \mid y}$ and $q_{\rm f \mid y}$ denote the pushforwards of $p_{\rm x \mid y}$ and $q_{\rm x \mid y}$ conditioned on $y$ and $y'$, respectively: $f \#p_{\rm x \mid y}(x \mid y)$, $f \#q_{\rm x \mid y}(x \mid y')$. The couplings $\gamma_{\rm f}^{\star}$ and $\gamma_{y}^{\star}$ are optimal transport plans between pairs $f\#p_{\rm x}$, $f\#q_{\rm x}$ and $p_y$, $q_y$.        
\end{definition}

By combining Lemmas~\ref{lemma1} and~\ref{lemma2}, letting $\alpha=1$ and defining a decomposable cost function with $c_1 = c_2 = \ell$, we get the following corollary.

\begin{corollary}\label{cor0}
    Suppose the loss function $\ell \colon \mathcal{Y} \times \mathcal{Y}
    \rightarrow \mathbb{R}$ satisfies Assumption~\ref{assum:metric} and a classifier $f \colon \mathcal{X} \rightarrow \mathcal{Y}$ satisfies Assumption~\ref{assum:surjective} then
    \begin{align}
        \risk_{q}(f) &\leq \risk_{p}(f) + W_{1, \ell}(f\#p_{\rm x}, f\#q_{\rm x}) 
        +
        \labelentangle(f)
        \label{eq:bound_in_x}, \\
        \risk_{q}(f) &\leq \risk_{p}(f) + W_{1, \ell}(p_{\rm y}, q_{\rm y}) 
        +
        \predentangle(f).
        \label{eq:bound_in_y}
    \end{align}
\end{corollary}

From the corollary we can see clearly the implications of aligning marginals during UDA. For instance, in the case of an optimal transport based alignment, learning a hypothesis $f$ that brings the set of pairs $f(x)$ and $f(x')$ closer (thus minimizing the Wasserstein distance between $f\#p_{\rm x}$ and $f\#q_{\rm x}$) does not ensure that the \emph{average} distance (with respect to the optimal coupling) between the conditionals $f \# p_{\rm y \mid x}$ and $f \# q_{\rm y \mid x}$ will also decrease. We refer to this potential increase in average conditional distance as \emph{entanglement}\footnote{Note the relation to the notion of spurious features in the out-of-domain generalization (OOD) literature~\citep{arjovsky2019invariant}: entanglement is the corresponding notion for optimal transport.}.

\subsection{Oracle Bound}

As discussed above, during the optimal transport association process, the source inputs $x$ can be associated to target inputs $x'$ that have different matching labels. Minimizing the marginal Wasserstein distance between such \emph{entangled} pairs (without including any entanglement terms from Definition~\ref{def:entangle}) can cause the entanglement terms $\labelentangle$ and $\predentangle$ to increase, potentially increasing the target risk as well. In order to better understand the roles that these two terms play, we will next relate them to each other. For this purpose, we first introduce an \emph{oracle} that we assume has access to the label entanglement $\labelentangle$.
\begin{definition}
    \label{def:oracle-upper-bound}
    We denote the \emph{oracle upper bound} (OUB) as
    \begin{equation}
        \oeub(f) \defeq 
        \risk_{p}(f) + W_{1, \ell}(f\#p_{\rm x}, f\#q_{\rm x}) 
        +
        \labelentangle(f).
    \end{equation}
\end{definition}
%
%
Although OUB includes only the label entanglement $\labelentangle(f)$, we note that both of the entanglement terms are intrinsically connected.
\begin{theorem}
    \label{thm:entangle-convert}
    Suppose Assumptions~\ref{assum:metric} and \ref{assum:surjective} hold. Then we have for all $f \in \mathcal{H}$,
    \begin{align*}
        \labelentangle(f) &\leq \predentangle(f) + \risk_p(f) + \risk_q(f) + W_{1, \ell}(p_{\rm y}, q_{\rm y}), \\
        \predentangle(f) &\leq \labelentangle(f) + \risk_p(f) + \risk_q(f) + W_{1, \ell}(f\# p_{\rm x}, f\# q_{\rm x}).
    \end{align*}
\end{theorem}
\removepf{
\begin{proof}[Proof Sketch]
    The proof is a result of applying triangle inequality to the loss function used to compute the entanglement terms, and then noting that each term corresponds to the source risk, target risk and the marginal Wasserstein distance respectively. The marginal Wasserstein distances can further be upper bounded by switching between the lower and the upper bounds in Lemma \ref{lemma2}.
\end{proof}
}
As a consequence, minimization of the bounds \eqref{eq:bound_in_x} and \eqref{eq:bound_in_y} is also connected together.
\begin{corollary}
    \label{cor:convert-objective}
    Suppose Assumptions~\ref{assum:metric} and~\ref{assum:surjective} hold. Then we have for all $f \in \mathcal{H}$,
    %
    %
    \begin{align*}
        \risk_{p}(f) + W_{1, \ell}(p_{\rm y}, q_{\rm y}) 
        +
        \predentangle(f) \in \left[ \frac{1}{3} \cdot \oeub(f), 3 \cdot \oeub(f) \right].
    \end{align*}
\end{corollary}
A consequence of Corollary~\ref{cor:convert-objective} is that minimizing either of the bounds \eqref{eq:bound_in_x} or \eqref{eq:bound_in_y} ($\mathcal{U}(f)$ corresponding to \eqref{eq:bound_in_x}) results in the minimization of the other.

%% file: sections/analysis.tex
\section{Analysis of Entanglement}
\label{sec:entanglement}

We now study the bounds in Corollary~\ref{cor0} carefully and discuss how we can use them to analyze UDA and the unoptimizable entanglement term. In particular, we will focus on (a) assumptions necessary for further analysis, (b) effect of label shift, and (c) bounding the effect of the unoptimizable (without target labels) entanglement term $\labelentangle(f)$ appearing in the oracle bound $\oeub(f)$.
%

\subsection{Assumptions}

In domain adaptation, a key quantity used to analyse the limit performance of a set of hypotheses $\mathcal{H}$ is the \emph{ideal joint error}~\citep{benDavid2010}. In this paper, we will refer to it as \emph{Low Joint Error}.


\begin{assumption}
    The \emph{Low Joint Error} assumption $\LJE(\lambda)$ holds for $\lambda > 0$ and a hypothesis class $\mathcal{H}$ if $\min_{f \in \mathcal{H}} \left\{ \risk_{p}(f) + \risk_{q}(f) \right\} < \lambda$.
\end{assumption}

The $\LJE(\lambda)$ assumption states the existence of a classifier $f^{\star}_{\LJE} \in \mathcal{H}$ which has low risk on both the source and the target distributions.
Without this assumption, we cannot hope to find a hypothesis with low source risk that transfers well to the target distribution (which is the goal of UDA as stated above).

We consider an alternative assumption which helps us control the label entanglement of a hypothesis set.

\begin{assumption}
    The \emph{Close Conditionals Assumption} $\CCA(\kappa)$ holds for $\kappa > 0$ and a hypothesis class $\mathcal{H}$ if there exists a classifier $f \in \mathcal{H}$ such that $\risk_{p}(f) + \max_{y \in \mathcal{Y}} W_{1, \ell}(f\# p_{\rm x \mid \rm y}(\cdot \mid y), f\# q_{\rm x \mid \rm y}(\cdot \mid y)) < \kappa$.
\end{assumption}


Instead of requiring low risk on both the source and target distributions, $\CCA(\kappa)$ requires a low source risk hypothesis $f^\star_{\CCA} \in \mathcal{H}$ which \emph{pushes} the conditionals close to each other: i.e, the assumption ensures that the pushforwards of conditionals $f^\star_{\CCA} \# p_{\rm x \mid y}(\cdot \mid y)$ and $f^\star_{\CCA} \# q_{\rm x \mid y}(\cdot \mid y)$ do not differ too much when conditioned on the \emph{same} label $y$.
$\CCA$ is generally a valid assumption for DNNs with large enough capacity. We discuss this property of DNNs in Appendix \ref{add_remarks} in detail.

One property of the $\CCA(\kappa)$ assumption is that its violation (for a large enough $\kappa > 0$) precludes certain hypotheses from achieving good target risk.

\begin{lemma}
    \label{lem:not-cca}
    Suppose Assumption~\ref{assum:metric} holds.
    If $\CCA(\kappa)$ does not hold, then for all $f \in \mathcal{H}$,
    \begin{equation*}
        \kappa \cdot q_{\rm y}(y_{\rm min}) \leq \risk_p(f) \cdot q_{\rm y}(y_{\rm max}) +\risk_q(f) + \risk_{r}(f),
    \end{equation*}
    where $r(x, y) = p_{\rm x \mid y}(x \mid y) \cdot q_{\rm y}(y) \in \triangle(\mathcal{X} \times \mathcal{Y})$ and $y_{\rm min} = \arg \min q_{\rm y}(y)$, $y_{\rm max} = \arg \max q_{\rm y}(y)$.
\end{lemma}


From this lemma we can infer that whenever $\CCA(\kappa)$ is violated (i.e., $\kappa$ is large) while $\LJE(\lambda)$ holds for some small $\lambda$, then either there is a large imbalance in the target label probabilities $q_{\rm y}$ or there must be a large discrepancy in the components of source risk for different labels. The latter case highlights the fact that in addition to low $\LJE(\lambda)$, we may also need a \emph{balanced} (among minority classes) source risk for $\CCA(\kappa)$ to hold for low $\kappa$. Consider the following example to illustrate this point further.





\paragraph{Example.}
Consider the setting of binary classification where $\mathcal{Y} = \{ 0, 1 \}$ and $q_{\rm y}(y) = 1/2$ for all $y \in \mathcal{Y}$. 
Now suppose that $\LJE(\lambda)$ holds and $ \CCA(\kappa) $ is violated for $\lambda, \kappa$ satisfying $C \leq \kappa / 2 - \lambda$ for some large $C > 0$. In this case, for any model achieving the LJE assumption $f^{\star}_{\LJE} \in \mathcal{H}$, Lemma~\ref{lem:not-cca} can be weakened to $ C \leq \risk_r(f^\star_\LJE)$. 
From $\LJE(\lambda)$ we have $\risk_p(f^\star_\LJE) < \lambda$. Thus, for this to be true, we require $\expect_{p}[\ell(f^\star_\LJE(x), y) \mid y = 0]$ to differ significantly from $\expect_{p}[\ell(f^\star_\LJE(x), y) \mid y = 1]$. In particular, the performance on the minority label $y_{\rm min} = \arg\min_y p_{\rm y}(y)$ on the source will be poor, \ie, $\expect_{p}[\ell(f^\star_\LJE(x), y) \mid y = y_{\rm min}]$ large.

\subsection{Effect of Label Shift}
In the following, we explore the effect of label shift between the source and target distributions $p_{\rm y}$ and $q_{\rm y}$.
%
\begin{lemma}\label{lem:close-marginals-cda}
    Suppose that Assumptions~\ref{assum:metric} to~\ref{assum:surjective} hold and there exists an $l > 0$ such that $\min_{y \neq y'} \ell(y, y') > l$. Then given $W_{1,\ell}(p_y, q_y) < \delta$ holds for some $\delta > 0$ and $\CCA(\kappa)$ for some $\kappa > 0$ then $\LJE(\lambda)$ holds with $\lambda = 2\kappa + \frac{L + l}{l} \cdot \delta$.
\end{lemma}
\removepf{
\begin{proof}[Proof Sketch]
    The proof is a consequence of the fact that if the source and target label distributions are close together in Wasserstein distance then the corresponding coupling is concentrated on the diagonals: off diagonal terms are small and by using upper and lower bounds of the loss, we can cap their effect on the target risk using equation~\eqref{eq:bound_in_y}.
\end{proof}
}
The above Lemma~\ref{lem:close-marginals-cda} shows the heavy price we might have to pay (in terms of loss) for small deviations of the label distribution $p(y)$: small deviations $\delta$, measured in terms of Wasserstein distance, can be amplified up to $L/l$ times. If the model $f$ outputs probability distributions (i.e., values in the simplex), then we can use the diameter of the simplex as measured by the metric $\ell(f, f')$ as the bound $L$. For the lower bound $l$, we note that for discrete $y, y'$ in the simplex $\min_{y \neq y'}\ell(y,y')$ can be seen as a scaling factor in the bounds.

Further note that if the marginal label distributions in the source and target domains are the same, then the LJE and CC assumptions provide a guarantee for each other.

\begin{corollary}
    \label{cor:equal-label-marginals-assum}
    Suppose Assumptions~\ref{assum:metric} to~\ref{assum:surjective} hold and $p_{\rm y} = q_{\rm y}$.
    Then $\CCA(\kappa)$ implies $\LJE(2\kappa)$ and 
    $\LJE(\lambda)$ implies $\CCA(\lambda \cdot (1 + q_{\rm y}(y_{\rm max})) / q_{\rm y}(y_{\rm min}))$.
\end{corollary}


\subsection{Optimization Criteria}
%
As the entanglement computation requires target labels, we also define the objective function without the entanglement term below, as a regularized source risk, which is the quantity that is usually minimized in practice~\citep{courty2017, shen2018wasserstein}. 
\begin{definition}
    \label{def:w-regularized-risk}
    We denote the Wasserstein Regularized Risk (WRR) as
    \begin{equation}
        \wrr(f) \defeq 
        \risk_{p}(f) + W_{1, \ell}(f\#p_{\rm x}, f\#q_{\rm x}).
    \end{equation}
\end{definition}
Based on the bound in \eqref{eq:bound_in_x} we analyze below the properties of the oracle entanglement upper bound $\oeub(f)$~\eqref{def:oracle-upper-bound} and the corresponding $\wrr(f)$~\eqref{def:w-regularized-risk}. We first start with a lower bound on $\oeub(f)$.
\begin{lemma}\label{lem:lower_bound_usr}
    Suppose Assumptions~\ref{assum:metric} to~\ref{assum:surjective} hold. Then
    $W_{1,\ell}(p_{\rm y}, q_{\rm y}) \leq \oeub(f)$.
\end{lemma}

\begin{proof}
    The proof follows from using the second lower bound and then the first upper bound in Lemma~\ref{lemma2}.
\end{proof}
Lemma~\ref{lem:lower_bound_usr} implies that although there may be a hypothesis $f_{\CCA}^{*}$ satisfying $\CCA(\kappa)$ for a small $\kappa > 0$, if there is a shift in the label distributions, an algorithm minimizing $\oeub(f)$ is not able to recover such a hypothesis.\footnote{Note that similar lower bounds were presented in~\citet{zhao2019learning} using Jensen-Shannon divergence.} 
We now have what we need to analyze the tightness of the bounds in \eqref{eq:bound_in_x}. We will show that $\CCA(\kappa)$ and low label shift implies that the bound in \eqref{eq:bound_in_x} is tight (up to small multiples of $\kappa$ and potentially large multiples of $\delta$) at the best hypothesis $f_\CCA^{\star}$.
\begin{lemma}
    \label{cor:oeub-cca-upperbound}
    Suppose that Assumptions~\ref{assum:metric} to~\ref{assum:surjective} hold, $W_{1,\ell}(p_{\rm y}, q_{\rm y}) < \delta$, and there exists an $l > 0$ such that $\min_{y \neq y'} \ell(y, y') > l$. If $f_\CCA^\star \in \mathcal{H}$ witnesses $\CCA(\kappa)$, then
    \begin{equation*}
        \oeub(f_\CCA^\star) \leq 3 \cdot \left( \kappa + \frac{L + l}{l} \cdot \delta \right).
    \end{equation*}
\end{lemma}
%
The above lemma implies that we can minimize $\oeub(f)$ to obtain small target risk if the label shift is small, assuming that we can find its minimizer. However, although the bound can be \emph{tight} at the best hypothesis $f_{\CCA}^{\star}$ satisfying $\CCA(\kappa)$, 
the entanglement term may not be small at the minimizer of $\wrr(f)$. We next look at further assumptions we can make to force this term to be smaller. 

\begin{assumption}[Gradual Shift Assumption]
    The Gradual Shift assumption $\GS(a, b, \varepsilon, s)$ holds for a classifier $f \in \mathcal{H}$ if $ \risk_p(f) < b $ implies that there exists $q^{(i)}_{\rm x \mid y}(\cdot \mid y) \in \triangle(\mathcal{X})$ such that
    \begin{equation*}
       q_{\rm x \mid \rm y} (x \mid y) = \sum_{i=1}^{s} r_i q^{(i)}_{\rm x \mid y}(x \mid y)
    \end{equation*}
    for $0 \leq r_i \leq a$ and all $y \in \mathcal{Y}$; and for all $i \in \{1, \ldots, s \}$
    we have
    $
        W_{1, \ell}(f\# q^{(i-1)}_{\rm x \mid y}(\cdot \mid y), f\# q^{(i)}_{\rm x \mid y}(\cdot \mid y)) < \varepsilon
    $,
    where $q^{(0)}_{\rm x \mid y}(x \mid y) \defeq p_{\rm x \mid y}(x \mid y)$.
\end{assumption}

Note that this definition is different from the usual gradual shift defined in the literature~\citep{kumar2020understanding}, in that the indices of the gradually shifting distributions may not be available, and the samples from the mixture components can be scrambled. Moreover note the dependency on $b$: besides the smoothness assumption on the shifting conditionals, we are also implicitly assuming regularity properties on $\mathcal{H}$\footnote{For example, if all hypotheses in $\mathcal{H}$ require the use of certain features to achieve low risk $b$, then we can relate the assumption further to the utilization of such \emph{stable} features~\citep{arjovsky2019invariant}.}

We first connect the $\GS$ assumption to that of $\CCA$.

\begin{lemma}
    \label{lem:gs-cca}
    Suppose that Assumptions~\ref{assum:metric} and $\GS(a, b, \varepsilon, s)$ hold.
    For a classifier $f \in \mathcal{H}$, if $\risk_p(f) < b$, then $\CCA(\kappa)$ holds with 
    $\kappa = b + \varepsilon \cdot \tfrac{a}{2}s (s+1)$.
\end{lemma}

\begin{proof}[Proof Sketch]
    Note that the $\GS$ assumption holds uniformly for each $y \in \mathcal{Y}$. The proof follows directly from the convexity of Wasserstein distances (with respect to their arguments) and the triangle inequality applied to the Wasserstein distances between the conditional source distributions $p_{\rm x \mid y}$ and conditional target distributions $q_{\rm x \mid y^{(i)}}$, for each $i = 1, \ldots, s$.
\end{proof}

Given Lemma~\ref{lem:gs-cca}, we can guarantee the $\CCA$ assumption whenever the GS assumption holds. Although the quadratic dependence on $s$ seems strong, one can examine certain scenarios where it is weakened. First note that $\varepsilon' = s \varepsilon$ intuitively can be seen as the full aggregated distance of the source to target distributions. Thus fixing $\varepsilon'$, increasing $s$ only provides a linear increase. Secondly, the mixture constraint is bounded as $a \in [1/s, 1]$. In the best case scenario, we have a balanced mixture $a = 1/s$. Then the dependence can be simplified as $\bigoh(b + a \varepsilon' s) = \bigoh(b + \varepsilon')$.



Finally we prove that the $\GS$ assumption entails slowly growing (depending on $s$) label entanglement for small label shift.

\begin{theorem}
    \label{thm:predentangle-upper}
    Suppose that Assumptions~\ref{assum:metric} to \ref{assum:surjective} and $\GS(a, b, \varepsilon, s)$ hold; $W_{1,l}(p_y, q_y) < \delta$, and there exists an $l > 0$ such that $\min_{y \neq y'} \ell(y, y') > l$. For a classifier $f \in \mathcal{H}$, if $\risk_p(f) < b$, then
    \snegativespace
    \begin{align*}
        \labelentangle(f) \leq 2b &+ \varepsilon \cdot as(s+1) + 2\delta \cdot \frac{L + l}{l}.
    \end{align*}
\end{theorem}

\begin{proof}[Proof Sketch]
    We first bound the prediction entanglement $\predentangle(f)$ using Lemma \ref{lem:gs-cca} and $\delta$, following the same steps used to prove Lemma \ref{lem:close-marginals-cda}. The label entanglement $\labelentangle(f)$ can then be bounded using Theorem \ref{thm:entangle-convert}.
\end{proof}

Theorem \ref{thm:predentangle-upper} states that under the previous assumptions, the label entanglement of a good source hypothesis ($\risk_p(f) < b$ for some small $b$) can be controlled. An immediate observation is that when $\delta = 0$, the label entanglement $\labelentangle(f)$ is upper bounded with the same dependencies as CCA, as per Lemma~\ref{lem:gs-cca}.

%% file: sections/experiments.tex
\section{Experiments}\label{sec:experiments}


In this section we experiment with several different models in different domain shift \emph{scenarios}, and show the wide variety of behaviours we can observe from different (model, optimizer, dataset) triplets: domain matching algorithms optimizing only for marginal alignment are shown to be susceptible to entanglement, and this vulnerability cannot be prevented purely through algorithmic means (such as hyperparameter tuning). Moreover we evaluate the oracles $\LJE$ and $\CCA$ to show the limit of what we can expect from the algorithms: if $\LJE$ holds for a small $\lambda$ but $\CCA(\kappa)$ does not, then we show that they do not perform well in general\footnote{Scripts used to generate the experiment results are available in \url{https://github.com/okankoc/Domain-Adaptation-and-Entanglement}.}.

\begin{table*}[h!]
    \centering
    \small
    \caption{%
        Performance of different algorithms including the LJE oracle objective minimization tested over various distribution shift scenarios. Each cell reports ``source accuracy / target accuracy'' and approaches with positive transfer (\ie, better than ERM) are notated by an underline \underline{x} and the best overall method is notated by a box \fbox{y}.
        (Top table) presents the best model \emph{conv1} and (Bottom table) presents a selection of other model results.
        }    
    \vspace{0.1cm}
    \begin{tabular}{lccccc}
    \toprule
     Scenario + Best Model & LJE Oracle & ERM & WRR & JDOT & DANN \\
     \midrule
     MNIST $\to$ USPS / conv1 & 97.6 / 94.7 & 98.0 / 83.8 & 96.9 / \underline{86.0} & 98.2 / \underline{87.1} & 96.2 / \fbox{89.3} \\
     USPS $\to$ MNIST / conv1 & 95.2 / 98.0 & 94.8 / 71.4 & 94.8 / \underline{78.8} & 95.6 / \fbox{89.3} &  93.3 / \underline{77.9} \\
     MNIST $\to$ MNIST-M / conv1 & 99.2 / 96.8 & 98.9 / 47.5 & 98.8 / \underline{51.2} & 99.1 / \underline{69.3} & 98.8 / \fbox{79.1} \\
     SVHN $\to$ MNIST / conv1 & 92.3 / 99.1 & 91.2 / 64.9 & 90.4 / \fbox{65.3} & 90.4 / 48.0 & 86.6 / 10.3 \\
     Portraits $1905\text{-}1974 \to 1974\text{-}2013$ / conv1 & 96.9 / 95.5 & 96.8 / 86.3 & 94.8 / 83.8 & 96.5 / 85.2 & 95.8 / \fbox{87.3} \\
     CIFAR10 $\to$ CIFAR10c / conv1 & 72.7 / 83.8 & 75.5 / 37.7 & 71.2 / 10.0 & 10.0 / 10.0 & 73.7 / \fbox{60.9} \\
     \bottomrule
    \end{tabular}

    \vspace{0.5cm}
    
    \begin{tabular}{lccccc}
    \toprule
     Scenario + Other Models & LJE Oracle & ERM & WRR & JDOT & DANN \\
     \midrule
     MNIST $\to$ USPS / MLP & 92.4 / 91.4 & 92.6 / 16.0 & 90.8 / \fbox{59.3} & 92.4 / \underline{37.8} & 91.4 / \underline{28.8} \\
     USPS $\to$ MNIST / MLP & 92.7 / 94.7 & 91.8 / 29.0 & 90.3 / \fbox{52.1} & 92.3 / \underline{50.5} & 91.6 / \underline{43.1} \\
     MNIST $\to$ MNIST-M / LeNet & 98.7 / 94.5 & 98.9 / 48.4 & 93.6 / \underline{56.3} & 99.0 / \fbox{58.5} & 89.6 / 19.9 \\
     SVHN $\to$ MNIST / conv2 & 90.9 / 98.9 & 90.1 / 64.4 & 19.6 / 11.3 & 90.6 / \fbox{68.7} & 80.6 / 18.7 \\
     Portraits $1905\text{-}1974 \to 1974\text{-}2013$ / MLP & 94.8 / 90.5 & 94.5 / 83.5 & 90.9 / 77.9 & 94.6 / \fbox{84.1} & 94.6 / 83.0 \\
     CIFAR10 $\to$ CIFAR10c / small$\_$cnn & 67.4 / 88.0 & 70.3 / 39.6 & 58.2 / 10.0 & 10.0 / 10.1 & 68.3 / \fbox{52.7} \\
     \bottomrule
    \end{tabular}
   %
   %
   %
   %
   %
    \label{table1}
\end{table*}


We report in Table~\ref{table1} mean source and target accuracies of various oracle and non-oracle UDA algorithms after $10$ epochs, evaluated over three different initializations. All methods in Table~\ref{table1} use cross-entropy as the loss function, the Adam optimizer with a learning rate of $0.001$ and a batchsize of $64$. We test with five different models of various depths, using fully-connected and convolutional networks. See Appendix~\ref{sec:experiment_details} for detailed explanations of the models, scenarios and the algorithms used. Different ablations of various parameters and settings are conducted in Appendix~\ref{app:ablations}, including batchsize variations and other optimizer configurations. We also consider in that section a more difficult dataset together with a bigger model, and add two more UDA algorithms.


%
\subsection{Results and Discussion}
Experiments in Table~\ref{table1} show overall that the vanilla Wasserstein distance regularized source risk minimization (WRR) in \eqref{def:w-regularized-risk} that we analyze in this paper is not worse off than other methods in general. 
We see that it has positive transfer (i.e., the target accuracy is higher than that of ERM) using the best model (top table) in the first four scenarios and for the other models tested (bottom table) in the first three scenarios. More importantly, when minimizing $\wrr(f)$, experiments with different models show very different target accuracies resulting from various degrees of entanglement\footnote{See Appendix~\ref{sec:experiment_details} for detailed accuracy plots supporting this conclusion.}. Depending on the model, even when the $\LJE$ assumption is satisfied, target risk varies significantly between cases of highly positive transfer to highly negative transfer. In Table~\ref{table1} we underline all the positive transfer cases and among them box the best performing method.

\begin{table*}[t!]
  \centering
  \small
    \caption{Comparison of the accuracy of minimizing LJE and $\CCA$ with quantities associated with the WRR objective. Cells formatted as X / Y correspond to the metric evaluated on the source / target data. $W_1(\hat{p}_y, \hat{q}_y)$ corresponds to the empirical Wasserstein distance between the marginal label distributions. The values under the ``Oracle Accuracy" header correspond to the accuracy values. The values under ``WRR" correspond to empirical estimates of the accuracy, risk, WRR objective and label-entanglement, respectively.}
    \vspace{0.1cm}
  \scalebox{0.95}{  
  \begin{tabular}{lccccccc}
    \toprule
    \multirow{2}*{Scenario + Model} & \multirow{2}*{$W_{1}(\hat{p}_{\rm y}, \hat{q}_{\rm y})$} & \multicolumn{2}{c}{Oracle Accuracy} & \multicolumn{4}{c}{WRR} \\
    \cmidrule(lr){3-4}
    \cmidrule(lr){5-8}
    & & $\LJE$ & $\CCA$ & Acc. & $\hat{\risk}_p$ / $\hat{\risk}_q$ & $\hat{\wrr}$ & $\hat{\labelentangle}$ \\
    \midrule
    MNIST $\to$ USPS / conv1 & 0.13 & 97.6 / 94.7 & 95.3 / 93.0 & 98.5 / 91.7 & 0.06 / 0.15 & 0.23 & 0.08 \\
    MNIST $\to$ USPS / MLP & 0.13 & 92.4 / 91.4 & 93.1 / 87.9 & 95.6 / 74.3 & 0.14 / 0.42 & 0.37 & 0.19 \\
    MNIST $\to$ MNIST-M / conv1  & 0.00 & 99.2 / 96.8 & 98.6 / 95.3 & 98.8 / 70.7 & 0.02 / 0.42 & 0.19 & 0.30 \\
    SVHN $\to$ MNIST / conv2 & 0.20 & 90.9 / 98.9 & 86.8 / 93.9 & 89.1 / 64.8 & 0.19 / 0.50 & 0.38 & 0.34 \\
    Portraits $1905\text{-}1974 \to 1974\text{-}2013$ MLP & 0.05 & 94.8 / 90.5 & 91.6 / 87.3 & 94.0 / 79.6 & 0.09 / 0.29 & 0.11 & 0.33 \\
    CIFAR10 $\to$ CIFAR10c / conv1 & 0.00 & 75.2 / 87.2 & 63.7 / 9.9 & 73.9 / 9.7 & 0.48 / 1.13 & 0.77 & 0.58 \\
    \bottomrule
  \end{tabular}
  }
  \label{table2}
\end{table*}

\paragraph{Oracle Minimization and Entanglement.} Next we select WRR for various extended experiments in Table~\ref{table2} to get a more quantitative understanding into our assumptions underlying marginal distance minimization and the resulting entanglement estimates. We use the oracles to understand better what is happening during marginal alignment: we optimize not only the LJE oracle but also the $\CCA$ oracle objective functions, as well as reporting estimates of the label entanglement $\hat{\labelentangle}(f_{\wrr})$. Unlike Table~\ref{table1}, we use the Euclidean distance as the loss function to be able to compare the values of each component (source risk, the marginal Wasserstein distance and the entanglement estimate). 
Incidentally, we see that using the Euclidean distance as the loss function outperforms using cross-entropy (this is another ablation setting which we explore further in Appendix~\ref{sec:experiment_details}). 


We indicate some interesting findings from Table~\ref{table2}. For the CIFAR-10c corruptions scenario, the $\CCA$-oracle minimization is unable to find a good hypothesis that will generalize to both environments, even though the LJE assumption is satisfied for a relatively small $\lambda$ (and in spite of the fact that label shift is zero for this scenario). This potentially explains why the target accuracy of WRR is so low for this dataset. Likewise, the estimates of label shift shown in the second column correlate with the relatively poorer performances of WRR, \eg, in the SVHN-MNIST scenario where we computed a label shift estimate of $W_1(\hat{p}_y, \hat{q}_y) = 0.2$, the highest shift among all tested scenarios. In both cases (high $\kappa$ estimate and high label shift estimates) the entanglement estimates $\hat{\labelentangle}(f_{\rm WRR})$ as well as the value of $\hat{\wrr}(f_{\rm WRR})$ are quite high, indicating that besides entanglement, the WRR minimization is also unable to find minima with suitably low marginal distances.





%% file: sections/conclusion.tex
\section{Conclusion}

In unsupervised domain adaptation (UDA), one common approach involves minimizing the source risk with a dissimilarity based regularizer that aligns marginal distributions. In the case of Wasserstein distances, we showed that regularizing the source risk by the marginal distance is not sufficient to get good target performance in general. Indeed, our bounds in Corollary~\ref{cor0} elucidate unoptimizable terms -- such as the \emph{label entanglement}, which may not decrease even when the source risk and marginal distance both decrease. This entanglement term intuitively corresponds to the average effect of not using \emph{invariant features}~\citep{arjovsky2019invariant, zhao2019learning}. 
Through our analysis, we showed that the assumptions such as the Close Conditionals ($\CCA$) or Gradual Shift (GS) assumptions can lead to stronger theoretical guarantees, limiting the effect of the entanglement. In addition, experimentally we observed that the entanglement term is a good tool for analyzing the optimization landscape of UDA algorithms: depending on the (model, optimizer, dataset) triplet we showed that the performance of UDA algorithms can vary significantly from cases of positive transfer to cases of highly negative transfer and that target accuracies correlate strongly with the degree of entanglement. Future work could include analyzing the optimization process more closely or exploring other reasonable assumptions to make UDA more tractable.

%% file: appendix.tex
\appendix

\section*{Appendix}

\input{appendix/copula}
\input{appendix/proofs_section_three}
\input{appendix/proofs_section_four}
\input{appendix/related_work}
\input{appendix/approximate-triangle-inequality}
\input{appendix/remarks}
\input{appendix/experiment_details}
\input{appendix/ablations}

%% file: appendix/copula.tex
\section{Coupling Properties}
\paragraph{Notation.} In the proofs shown in the Appendix, for convenience we group the variables integrated over, e.g., $\dd(y, y')$ indicates joint integration over the variables $y$ and $y'$.
We further utilize the Iverson brackets $\llbracket \texttt{predicate} \rrbracket$ which evaluate to $1$ if \texttt{predicate} is true and $0$ if it is false~\citep{knuth1992two}.

\begin{lemma}
    \label{lem:factorize-coupling}
    Suppose $\gamma_{\rm x} \in \Gamma(\mu_{\rm x}, \nu_{\rm x})$ and $\gamma_{\rm y \mid x}(\cdot \mid x, x') \in \Gamma(\mu_{\rm y \mid x}(\cdot \mid x), \nu_{\rm y \mid x}(\cdot \mid x'))$.
    Then for $\gamma = \gamma_{\rm x} \cdot \gamma_{\rm y \mid x}$, $\gamma \in \Gamma(\mu, \nu)$.
\end{lemma}
\begin{proof}
    Follows immediately from the calculation of marginals.
\end{proof}

\begin{lemma}\label{lem:transport-partial-decomp}
    Let $\mu, \nu \in \triangle(\mathcal{X} \times \mathcal{Y})$. Then for any coupling $\gamma \in \Gamma(\mu, \nu)$, there exists a factorization such that $\gamma = \gamma_{\rm x} \cdot \alpha$, where $\gamma_{\rm x} \in \Gamma(\mu_{\rm x}, \nu_{\rm x})$ and $\alpha(\cdot \mid x, x') \in \triangle(\mathcal{Y} \times \mathcal{Y})$ is some probability measure for each $x, x'$.

    Specifically, 
    \begin{equation*}
        \gamma_{\rm x}(x, x') = \int \gamma((x, y), (x', y')) \dd (y, y').
    \end{equation*}
\end{lemma}
\begin{proof}
    Define $\gamma_{\rm x}$ as per the lemma.
    One can then verify that marginalization yields $\mu_{\rm x}$ and $\nu_{\rm x}$. For example,
    \begin{align*}
        \int \gamma_{\rm x}(x, x') \dd x'
        &= \int \gamma((x, y), (x', y')) \dd (y, y') \dd x' \\
        &= \int \gamma((x, y), (x', y')) \dd (x', y') \dd y \\
        &= \int \mu(x, y) \dd y \\
        &= \mu_{\rm x}(x).
    \end{align*}
    The opposite marginalization follows identically. $\alpha$ is a (conditional) probability measure as a result of the above marginalization of $\gamma$. Note that $\alpha$ is not necessarily a coupling of conditionals, i.e., $\alpha(\cdot \mid x, x') \in \Gamma(\mu_{\rm y \mid x}(\cdot \mid x), \nu_{\rm y \mid x}(\cdot \mid x')$ may not hold.
\end{proof}

\begin{lemma}
    \label{lem:pushforward-coupling}
    Let $f \colon \mathcal{X} \rightarrow \mathcal{Y}$ be a measurable function and $\mu, \nu \in \triangle(\mathcal{X} \times \mathcal{Y})$. Suppose $\gamma \in \Gamma(\mu, \nu)$. 
    Then,
    \begin{equation}
        \bar{f}\#\gamma \in \Gamma(f\#\mu, f\#\nu).
    \end{equation}
    where $\bar{f} : (x, y, x', y') \to (f(x), y, f(x'), y')$.
\end{lemma}
\begin{proof}
    The result follows from simply utilizing the change of measure property of pushforward measures:
    \begin{align*}
        \int (\bar{f}\# \gamma)((\hat{y}, y), (\hat{y}', y')) \dd (\hat{y}', y')
        &= \int \gamma(f^{-1}(\hat{y}), y, f^{-1}(\hat{y}'), y') \dd (\hat{y}', y') \\
        &= \int \gamma(f^{-1}(\hat{y}), y, x', y') \dd (x', y') \\
        &= \mu(f^{-1}(\hat{y}), y) \\
        &= f\#\mu(\hat{y}, y).
    \end{align*}
    As the calculation holds symmetrically for marginalizing over the other argument, the result holds.
\end{proof}

%% file: appendix/proofs_section_three.tex
\section{Proofs}

\subsection{Proof of Lemma~\ref{lemma1}}

\begin{proof}
    To prove the result, for an arbitrary coupling $\gamma \in \Gamma(p, q)$ we consider the following:
    \begin{align*}
        \risk_{q}(f) &= \expect_{q}[\ell(f(x),y)] \\
        &= \expect_{p}[\ell(f(x),y)] + \expect_{q}[\ell(f(x),y)] - \expect_{p}[\ell(f(x),y)] \\
        &= \expect_{p}[\ell(f(x),y)] + \expect_{(x, y), (x', y') \sim \gamma}\left[ \ell(f(x'),y') - \ell(f(x),y) \right] \\
        &\leq \expect_{p}[\ell(f(x),y)] + \expect_{(x, y), (x', y') \sim \gamma}\left[ \ell(f(x), f(x')) + \ell(f(x), y) + \ell(y, y') - \ell(f(x),y) \right] \\
        &= \expect_{p}[\ell(f(x),y)] + \expect_{(x, y), (x', y') \sim \gamma}\left[ \ell(f(x), f(x')) + \ell(y, y')  \right],
    \end{align*}

    where the inequality holds due to our assumptions of symmetry and triangle inequality for $\ell$. This proves the first result after taking an infimum over couplings. For the next result, we need to further examine the expectation
    \begin{align*}
        \expect_{(x, y), (x', y') \sim \gamma}\left[ \ell(f(x), f(x')) + \ell(y, y')  \right]
        &= \int c_{\ell}(f(x), y, f(x'), y') \dd \gamma(x, y, x', y') \\
        &= \int c_{\ell}(\hat{y}, y, \hat{y}', y') \dd \bar{f}\#\gamma(\hat{y}, y, \hat{y}', y').
    \end{align*}
    From Lemma~\ref{lem:pushforward-coupling}, we have that $\bar{f}\# \gamma \in \Gamma(f\# p, f\# q)$ for $\bar{f}: (x, y, x', y') \to (f(x), y, f(x'), y')$. Now following \citet[Proposition 1.101]{leskela2010stochastic}, as $f$ is surjective, for any $\gamma' \in \Gamma(f\# p, f\# q)$ there exists a $\gamma \in \Gamma(p, q)$ such that the pushforward by $\bar{f}$ is equivalent to $\gamma'$.
    This implies that
    \begin{equation*}
        \Gamma(f\#p, f\#q) \subseteq \left\{ \bar{f}\#\gamma : \gamma \in \Gamma(p, q) \right\}.
    \end{equation*}
    
    As such, 
    the result follows after applying an infimum over couplings on the inequality.
\end{proof}

\subsection{Proof of Lemma~\ref{lemma2}}

To prove the result, we breakdown the proof into the cases of lower and upper bound.

%

\begin{proof}[Proof of Lower Bound]
    Let us denote $\gamma^{\star} \in \Gamma(\mu, \nu)$ as an optimal transport plan for $W_{1, c}(\mu, \nu)$. That is,
    \begin{align*}
        W_{1, c}(\mu, \nu)
        &= \iint c((x, y), (x', y')) \gamma^{\star}((x, y), (x', y')) \dd (y, y') \dd (x, x').
    \end{align*}
    Let $\gamma_{\rm x} \in \Gamma(\mu_{\rm x}, \nu_{\rm x})$ such that $\gamma^{\star} = \gamma_{\rm x} \cdot \alpha$, where $\alpha(\cdot \mid x, x') \in \triangle(\mathcal{Y} \times \mathcal{Y})$ for each $x, x'$, as per Lemma~\ref{lem:transport-partial-decomp}, 

    Thus we have
    \begin{align*}
        W^p_{p, c}(\mu, \nu)
        &= \iint c^p((x, y), (x', y')) \gamma^{\star}((x, y), (x', y')) \dd (y, y') \dd (x, x') \\
        &\geq \iint c^p_1(x, x')   \gamma^{\star}((x, y), (x', y')) \dd (y, y') \dd (x, x') \\
        &= \int c^p_1(x, x') \int \gamma^{\star}((x, y), (x', y')) \dd (y, y') \dd (x, x') \\
        &= \int c^p_1(x, x') \gamma_{\rm x}(x, x') \dd (x, x') \\
        &\geq \inf_{\gamma_{\rm x} \in \Gamma(\mu_{\rm x}, \nu_{\rm x})} \int c^p_1(x, x') \gamma_{\rm x}(x, x') \dd (x, x') \\
        &= W^p_{p, c_1}(\mu_{\rm x}, \nu_{\rm x}),
    \end{align*}
    where the first inequality comes from the non-negativity of cost functions: $c^p = (c_1 + c_2)^p \geq c_1^p$.

    The case for $W^p_{p, c_2}(\mu_{\rm y}, \nu_{\rm y})$ follows indentically.
\end{proof}

\begin{proof}[Proof of Upper Bound]
    We prove the first statement. The second inequality is proven identically, with $x$'s and $c_1$'s role exchanged with $y$'s and $c_2$'s. We start from the definition of the Wasserstein p-distance
    \begin{align*}
        W_{p, c}(\mu, \nu)
        &= \left( \inf_{\gamma \in \Gamma(\mu, \nu)} \left\{ \int c((x, y), (x', y'))^p \gamma((x, y), (x', y')) \dd (y, y') \dd (x, x') \right\} \right)^{1/p} \\
        &= \inf_{\gamma \in \Gamma(\mu, \nu)} \left \{ \int \left(c_1(x, x') + c_2(y, y')\right)^{p} \gamma((x, y), (x', y')) \dd (y, y') \dd (x, x') \right\}^{1/p} \\
        &\leq \inf_{\gamma \in \Gamma(\mu, \nu)} \left \{ \left( \int c^p_1(x, x') \gamma((x, y), (x', y')) \dd (y, y') \dd (x, x') \right)^{1/p} \right. \\
        &\quad \quad \left. + \left( \int c^p_2(y, y') \gamma((x, y), (x', y')) \dd (y, y') \dd (x, x') \right)^{1/p} \right \}
    \end{align*}
    where we applied Minkowski's inequality. We now define the sets,
    \begin{align*}
        \Gamma_{\rm x}(\mu, \nu) &= \Gamma(\mu_{x}, \nu_{x}) \\
        \Gamma_{\rm y \mid x}(\mu, \nu \mid x, x') &= \Gamma(\mu_{\rm y \mid x}(\cdot \mid x), \nu_{\rm y \mid x}(\cdot \mid x')) \\
        \Gamma^f(\mu, \nu) &= \{ \gamma \mid \gamma(x, y, x', y') = \gamma_{\rm x}(x, x') \cdot \gamma_{\rm y \mid x}(y, y' \mid x, x'), \\
        & \ \quad \quad \quad \gamma_{\rm x} \in \Gamma(\mu_{x}, \nu_{x}), \gamma_{\rm y \mid x}(\cdot \mid x, x') \in \Gamma_{\rm y \mid x}(\mu, \nu \mid x, x') \}.
    \end{align*}
    One can easily verify that $\Gamma^f(\mu, \nu) \subseteq \Gamma(\mu, \nu)$.
    Furthermore, let $\gamma_{\rm x}^{\star} = \mathop{\arg\inf}_{\gamma_{\rm x} \in \Gamma_{\rm x}(\mu, \nu)} \smash{\int} c^p_1(x, x') \dd {\gamma_{\rm x}} (x, x')$. As a result, we have
    \begin{align*}
        W_{p, c}(\mu, \nu)
        &\leq \inf_{\gamma \in \Gamma^f(\mu, \nu)} \left \{ \left( \int c^p_1(x, x') \gamma((x, y), (x', y')) \dd (y, y') \dd (x, x') \right)^{1/p} \right. \\
        &\quad \quad \left. + \left( \int c^p_2(y, y') \gamma((x, y), (x', y')) \dd (y, y') \dd (x, x') \right)^{1/p} \right \} \\
        &= \inf_{\gamma_{\rm x} \in \Gamma_{\rm x}(\mu, \nu)} \inf_{\gamma_{\rm y \mid x}(\cdot \mid x, x') \in \Gamma_{\rm y \mid x}(\mu, \nu \mid x, x')} \left \{ \left( \int c^p_1(x, x') \gamma_{\rm x}(x, x') \dd (x, x') \right)^{1/p} \right. \\
        & \quad \quad \quad \quad \quad \quad \left. 
        + \left( \int c^p_2(y, y') \gamma_{\rm x}(x, x')\gamma_{\rm y \mid x}(y, y' \mid x, x') \dd (y, y') \dd (x, x') \right)^{1/p} \right \} \\
        &\leq \inf_{\gamma_{\rm y \mid x}(\cdot \mid x, x') \in \Gamma_{\rm y \mid x}(\mu, \nu \mid x, x')} \left \{ \left( \int c^p_1(x, x') \gamma^{\star}_{\rm x}(x, x') \dd (x, x') \right)^{1/p} \right. \\
        & \quad \quad \quad \quad \quad \quad \left. 
        + \left( \int c^p_2(y, y') \gamma^{\star}_{\rm x}(x, x')\gamma_{\rm y \mid x}(y, y' \mid x, x') \dd (y, y') \dd (x, x') \right)^{1/p} \right \} \\
        &= \left( \int c^p_1(x, x') \gamma^{\star}_{\rm x}(x, x') \dd (x, x') \right)^{1/p} \\
        & \quad \quad \quad \quad 
        + \inf_{\gamma_{\rm y \mid x}(\cdot \mid x, x') \in \Gamma_{\rm y \mid x}(\mu, \nu \mid x, x')} \left \{ \left( \int c^p_2(y, y') \gamma^{\star}_{\rm x}(x, x')\gamma_{\rm y \mid x}(y, y' \mid x, x') \dd (y, y') \dd (x, x') \right)^{1/p} \right \} \\
        &= W_{p, c_1}(\mu_{\rm x}, \nu_{\rm x}) \\
        & \quad \quad \quad \quad 
        + \inf_{\gamma_{\rm y \mid x}(\cdot \mid x, x') \in \Gamma_{\rm y \mid x}(\mu, \nu \mid x, x')} \left \{ \left( \int c^p_2(y, y') \gamma^{\star}_{\rm x}(x, x')\gamma_{\rm y \mid x}(y, y' \mid x, x') \dd (y, y') \dd (x, x') \right)^{1/p} \right \} \\
        &= W_{p, c_1}(\mu_{\rm x}, \nu_{\rm x}) + \left( \int \gamma^{\star}_{\rm x}(x, x') \cdot W^p_{p, c_2}(\mu_{\rm y \mid x}(\cdot \mid x), \nu_{\rm y \mid x}(\cdot \mid x')) \dd (x, x') \right)^{1/p} \\
        &= W_{p, c_1}(\mu_{\rm x}, \nu_{\rm x}) + \expect_{\gamma_{x}^{\star}}\left[W_{p, c_2}^{p}(\mu_{\rm y \mid x}(\cdot \mid x), \nu_{\rm y \mid x}(\cdot \mid x')) \right]^{1/p}.
    \end{align*}
    (We note that the infimum $\inf_{\gamma_{\rm y \mid x}(\cdot \mid x, x') \in \Gamma_{\rm y \mid x}(\mu, \nu \mid x, x')}$ uses a slight abuse of notation where we are taking individual infimums over each $x, x'$ pairs.)
\end{proof}

\subsection{Proof of Corrollary~\ref{cor0}}

\begin{proof}
    Follows immediately from Lemma~\ref{lemma1} and~\ref{lemma2}.
\end{proof}

\subsection{Proof of Theorem~\ref{thm:entangle-convert}}

We first prove the following general Lemma.

\begin{lemma}
    \label{lem:lower_bound_on_combined_risk}
    Suppose that Assumption~\ref{assum:metric} holds. Then for all $f \in \mathcal{H}$
    \begin{equation*}
        \labelentangle(f)
        \leq 
        \risk_p(f) + \risk_q(f) + W_{1, \ell}(f\# p_{\rm x}, f\# q_{\rm x}).
    \end{equation*}
\end{lemma}
\begin{proof}
    Let $f \in \mathcal{H}$ be arbitrary. We begin by first upper bounding the label entanglement
    \begin{align*}
        \labelentangle(f) 
        &= \int W_{1, \ell}(p_{\rm y \mid f}(\cdot \mid \hat{y}), q_{\rm y \mid f}(\cdot \mid \hat{y}')) \dd \gamma_{\rm f}^\star(\hat{y}, \hat{y}') \\
        &= \iint \ell(y, y') \dd \gamma_{\rm y \mid f}(y, y' \mid \hat{y}, \hat{y}') \dd \gamma_{\rm f}^\star(\hat{y}, \hat{y}'),
    \end{align*}
    where $\gamma_{\rm f}^\star$ is the optimal transport plan between $f\#p_{\rm x}$ and $f\# q_{\rm x}$ and $\gamma_{\rm y \mid f}(\cdot \mid \hat{y}, \hat{y}')$ is the optimal transport plan between $p_{\rm y \mid f}$ and $q_{\rm y \mid f}$, with respect to their corresponding Wasserstein distances.

    Now, applying triangle inequality to the loss function $\ell$, assumed among the metric properties of $\ell$, we have
    \begin{align*}
        \labelentangle(f) 
        &= \iint \ell(y, y') \dd \gamma^\star_{\rm y \mid f}(y, y' \mid \hat{y}, \hat{y}') \dd \gamma_{\rm f}^\star(\hat{y}, \hat{y}') \\
        &\leq \iint \left( \ell(\hat{y}, y) + \ell(\hat{y}', y') + \ell(\hat{y}, \hat{y}') \right) \dd \gamma^\star_{\rm y \mid f}(y, y' \mid \hat{y}, \hat{y}') \dd \gamma_{\rm f}^\star(\hat{y}, \hat{y}') \\
        &= \risk_p(f) + \risk_q(f) + \iint \ell(\hat{y}, \hat{y}') \dd \gamma^\star_{\rm y \mid f}(y, y' \mid \hat{y}, \hat{y}') \dd \gamma_{\rm f}^\star(\hat{y}, \hat{y}') \\
        &= \risk_p(f) + \risk_q(f) + \int \ell(\hat{y}, \hat{y}') \dd \gamma_{\rm f}^\star(\hat{y}, \hat{y}') \\
        &= \risk_p(f) + \risk_q(f) + W_{1, \ell}(f\# p_{\rm x}, f\# q_{\rm x}),
    \end{align*}
    where the last equality follows from the optimality of the transportation plan.
\end{proof}

\begin{proof}[Proof of Theorem~\ref{thm:entangle-convert}]
    The proofs of the two inequalities are symmetric. As such we only provide a proof of the former label entanglement. From Lemma~\ref{lemma2}, we can switch from a lower bound to an upper bound to yield,
    \begin{equation*}
        W_{1, \ell}(f\# p_{\rm x}, f\# q_{\rm x}) 
        \leq 
        W_{1, \ell, \ell}(f\# p, f\# q)
        \leq W_{1, \ell}(p_{\rm y}, q_{\rm y}) + \predentangle(f).
    \end{equation*}
    Combining this fact together with Lemma~\ref{lem:lower_bound_on_combined_risk} completes the proof.
\end{proof}

\subsection{Proof of Corollary~\ref{cor:convert-objective}}

\begin{proof}
    We first prove the lower bound.
    Using Theorem~\ref{thm:entangle-convert}, we have
    \begin{align*}
        \oeub(f) \leq 2 \risk_{p}(f) + W_{1, \ell}(p_{\rm y}, q_{\rm y}) + \risk_{q}(f) + \predentangle(f) + W_{1, \ell}(f \# p_{\rm x}, f \# q_{\rm x}). 
    \end{align*}
    Now, we can upper bound $W_{1, \ell}(f \# p_{\rm x}, f \# q_{\rm x})$ using the same reasoning that was used in the proof of Theorem~\ref{thm:entangle-convert}, yielding
    \begin{align*}
        \oeub(f) 
        &\leq 2 \risk_{p}(f) + W_{1, \ell}(p_{\rm y}, q_{\rm y}) + \risk_{q}(f) + \predentangle(f) + W_{1, \ell}(f \# p_{\rm x}, f \# q_{\rm x}) \\
        &\leq 2 \risk_{p}(f) + 2W_{1, \ell}(p_{\rm y}, q_{\rm y}) + \risk_{q}(f) + 2\predentangle(f) \\
        &\leq 3 \risk_{p}(f) + 3W_{1, \ell}(p_{\rm y}, q_{\rm y}) + 3\predentangle(f).
    \end{align*}
    As required.

    The upper bound is proven identically, except by starting with $\risk_{p}(f) + W_{1, \ell}(p_{\rm y}, q_{\rm y}) + \predentangle(f)$ and switching the roles of $\predentangle(f)$ and $\labelentangle(f)$ in the proof above.
\end{proof}

%% file: appendix/proofs_section_four.tex
\subsection{Proof of Lemma~\ref{lem:not-cca}}

\begin{proof}
    As $\CCA(\kappa)$ does not hold, for all $f \in \mathcal{H}$, we have that
    \begin{equation*}
        \kappa < \risk_{p}(f) + \max_y W_{1, \ell}(f\# p_{\rm x \mid \rm y}(\cdot \mid y), f\# q_{\rm x \mid \rm y}(\cdot \mid y)).
    \end{equation*}
    Let us fix $f$ for now and denote $\bar{y}$ as the label of the conditional which satisfies the maximum.

    Now we have,
    \begin{align*}
        q_{\rm y}(\bar y) \cdot \kappa 
        &< q_{\rm y}(\bar y) \cdot \risk_{p}(f) + q_{\rm y}(\bar y) \cdot W_{1, \ell}(f\# p_{\rm x \mid \rm y}(\cdot \mid \bar{y}), f\# q_{\rm x \mid \rm y}(\cdot \mid \bar{y})) \\
        &\leq q_{\rm y}(\bar y) \cdot \risk_{p}(f) + \int q_{\rm y}(y) \cdot W_{1, \ell}(f\# p_{\rm x \mid \rm y}(\cdot \mid {y}), f\# q_{\rm x \mid \rm y}(\cdot \mid {y})) \dd y.
    \end{align*}
    Now denote $\gamma^{\star}(\hat{y}, \hat{y}' \mid y) $ as the optimal transport plan corresponding to each Wasserstein distance between the conditionals. Thus, we have
    \begin{align*}
        q_{\rm y}(\bar y) \cdot \kappa 
        &< q_{\rm y}(\bar y) \cdot \risk_{p}(f) + \int q_{\rm y}(y) \int \ell(\hat{y}, \hat{y}') \gamma^\star(\hat{y}, \hat{y}' \mid y) \dd(\hat{y}, \hat{y}') \dd y \\
        &\leq q_{\rm y}(\bar y) \cdot \risk_{p}(f) + \int q_{\rm y}(y) \int \left( \ell(\hat{y}, y) + \ell(\hat{y}', y)  \right) \gamma^\star(\hat{y}, \hat{y}' \mid y) \dd(\hat{y}, \hat{y}') \dd y \\
        &= q_{\rm y}(\bar y) \cdot \risk_{p}(f) + \int q_{\rm y}(y) \left( 
            \int \ell(\hat{y}, y) f\#p_{\rm x \mid y}(\hat{y}) \dd \hat{y}
            +
            \int \ell(\hat{y}', y) f\#q_{\rm x \mid y}(\hat{y}') \dd \hat{y}'
        \right) \dd y \\
        &= q_{\rm y}(\bar y) \cdot \risk_{p}(f) + \int q_{\rm y}(y) \left( 
            \int \ell(f(x), y) p_{\rm x \mid y}(x) \dd x
            +
            \int \ell(f(x'), y) q_{\rm x \mid y}(x') \dd x'
        \right) \dd y \\
        &= q_{\rm y}(\bar y) \cdot \risk_{p}(f) + \risk_{q}(f) + \risk_{p_{\rm x \mid y} q_{\rm y}}(f).
    \end{align*}
    The second inequality follows from the triangle inequality (among the assumed metric properties of $\ell$).
    
    Taking maximum and minimums to make the bound hold uniformly over $f \in \mathcal{H}$ yields the result.
\end{proof}

\subsection{Proof of Lemma~\ref{lem:close-marginals-cda}}

\begin{proof}
    Using the second upper bound in Lemma~\ref{lemma2} for 1-Wasserstein distances
    \begin{align*}
        \risk_q(f^\star) 
        &\leq \kappa + \delta + \int \llbracket{y \neq y'}\rrbracket W_{1,\ell}(f^\star \# p_{x \mid y}(\cdot \mid y), f^\star \# q_{x \mid y}(\cdot \mid y')) \dd\gamma_{\rm y}^{\star}(y, y'),
    \end{align*}
    where $f^\star$ witnesses our assumption of $\CCA(\kappa)$.

    Now all we need to do is to bound the terms on the conditionals. We utilize the boundedness assumption on $\ell$, yielding,
    \begin{align*}
        \risk_q(f^\star) 
        &\leq \kappa + \delta + L \cdot \int \llbracket{y \neq y'}\rrbracket \dd \gamma_{\rm y}^{\star}(y, y').
    \end{align*}

    Now the summation over the coupling can be bounded by lower bounding $W_{1,\ell}(p_{\rm y}, q_{\rm y})$,
    \begin{align*}
    l \int \llbracket{y \neq y'}\rrbracket \dd \gamma_{\rm y}^{\star}(y, y') 
    \leq 
    \int \ell(y, y') \dd \gamma_{y}^{\star}(y, y')
    <
     \delta 
     \implies
     \int \llbracket{y \neq y'}\rrbracket \dd \gamma_{\rm y}^{\star}(y, y') \leq \delta / l,
    \end{align*}
    where we use the assumption that $\ell(y, y') = 0$ when $y = y'$ and that $W_{1,\ell}(p_{\rm y}, q_{\rm y}) < \delta$.
    
    This yields,
    \begin{align*}
        \risk_q(f^\star) 
        &< \kappa + \delta + \frac{L}{l} \cdot \delta.
    \end{align*}

    Thus together, we have
    \begin{align*}
        \risk_p(f^\star) 
        +
        \risk_q(f^\star) 
        &< 2\kappa + \delta + \frac{L}{l} \cdot \delta.
    \end{align*}
    As required.
\end{proof}

\subsection{Proof of Corollary~\ref{cor:equal-label-marginals-assum}}

We split the proof of this corollary into two sub-results.

\begin{corollary}\label{cor42}
    Suppose Assumption~\ref{assum:metric} to~\ref{assum:surjective} holds and $p_{\rm y} = q_{\rm y}$.
    If $\CCA(\kappa)$ holds for $\kappa > 0$, then $\LJE(\lambda)$ holds for $\lambda = 2 \kappa$.
\end{corollary}
\begin{proof}
    Hold via Lemma~\ref{lem:close-marginals-cda} with $\delta = 0$.
\end{proof}

\begin{corollary}
    \label{cor:LJE-to-CCA}
    Suppose Assumption~\ref{assum:metric} holds and $p_{\rm y} = q_{\rm y}$. If $\LJE(\lambda)$ holds for $\lambda > 0$, then $\CCA(\kappa)$ holds for $\kappa = \lambda \cdot \frac{1 + \max_y q_{\rm y}(y)}{\min q_{\rm y}(y)}$.
\end{corollary}
\begin{proof}
    Holds via contra-positive of Lemma~\ref{lem:not-cca}.
\end{proof}

\begin{proof}[Proof of Corollary~\ref{cor:equal-label-marginals-assum}]
    Follows from Corollary~\ref{cor42} and~\ref{cor:LJE-to-CCA}.
\end{proof}

\subsection{Proof of Lemma~\ref{cor:oeub-cca-upperbound}}

\begin{proof}
    Let $f \in \mathcal{H}$ satisfy $\CCA(\kappa)$.
    Using Corollary~\ref{cor:convert-objective}, we have that
    \begin{align*}
        \oeub(f) 
        &\leq 3 \risk_p(f) + 3 W_{1,\ell}(p_{\rm y}, q_{\rm y}) + 3 \predentangle(f) \\
        &< 3 \delta + 3 \risk_p(f) + 3 \predentangle(f) \\
        &= 3 \delta + 3 \risk_p(f) + 3 \int_{\mathcal{Y} \times \mathcal{Y}} W_{1, \ell}(p_{\rm f \mid y}(\cdot \mid y), q_{\rm f \mid y}(\cdot \mid y')) \dd \gamma^\star_{\rm y}({y}, {y}') \\
        &= 3 \delta + 3 \risk_p(f) + 3 \int_{\mathcal{Y} \times \mathcal{Y}} \llbracket y = y' \rrbracket \cdot W_{1, \ell}(p_{\rm f \mid y}(\cdot \mid y), q_{\rm f \mid y}(\cdot \mid y')) \dd \gamma^\star_{\rm y}({y}, {y}') \\
        &\quad\quad\quad + 3 \int_{\mathcal{Y} \times \mathcal{Y}} \llbracket y \neq y' \rrbracket \cdot W_{1, \ell}(p_{\rm f \mid y}(\cdot \mid y), q_{\rm f \mid y}(\cdot \mid y')) \dd \gamma^\star_{\rm y}({y}, {y}') \\
        &< 3 \delta + 3 \kappa + 3 \int_{\mathcal{Y} \times \mathcal{Y}} \llbracket y \neq y' \rrbracket \cdot W_{1, \ell}(p_{\rm f \mid y}(\cdot \mid y), q_{\rm f \mid y}(\cdot \mid y')) \dd \gamma^\star_{\rm y}({y}, {y}') \\
        &\leq 3 \delta + 3 \kappa + 3L \int_{\mathcal{Y} \times \mathcal{Y}} \llbracket y \neq y' \rrbracket \dd \gamma^\star_{\rm y}({y}, {y}').
    \end{align*}
    Using the same argument as Lemma~\ref{lem:close-marginals-cda}, we have
    \begin{equation*}
        \int_{\mathcal{Y} \times \mathcal{Y}} \llbracket y \neq y' \rrbracket \dd \gamma^\star_{\rm y}({y}, {y}') \leq \delta / l.
    \end{equation*}
    The result follows immediately after.
\end{proof}

A simpler version of the lemma can also be proven:
\begin{corollary}
    \label{cor:oeub-upper-LJE}
    Suppose that Assumption~\ref{assum:metric}. Then $f \in \mathcal{H}$ satisfying $\LJE(\lambda)$, implies that
    \begin{equation*}
        \oeub(f) \leq 2\lambda + 2W_{1, \ell}(f\# p_{\rm x}, f\# q_{\rm x}).
    \end{equation*}
\end{corollary}
\begin{proof}
    The result yields immediately after upper bounding $\oeub(f)$ using the $\LJE(\lambda)$ assumption and Lemma~\ref{lem:lower_bound_on_combined_risk}.
\end{proof}

\subsection{Proof of Lemma~\ref{lem:gs-cca}}

\begin{proof}
    Let $f$ satisfy the requirements for GS.

    Following the convexity of Wasserstein distance, we have that
    \begin{align*}
        W_{1, \ell}(f\# p_{\rm x \mid y}(\cdot \mid y), f\# q_{\rm x \mid y}(\cdot \mid y))
        &\leq \sum_{i = 1}^s r_i \cdot  W_{1, \ell}(f\# p_{\rm x \mid y}(\cdot \mid y), f\# q^{(i)}_{\rm x \mid y}(\cdot \mid y)).
    \end{align*}
    Note that for each of $i$, applying triangle inequality
    \begin{align*}
        W_{1, \ell}(f\# p_{\rm x \mid y}(\cdot \mid y), f\# q^{(i)}_{\rm x \mid y}(\cdot \mid y))
        &\leq
        \sum_{j = 1}^{i}  W_{1, \ell}(f\# q^{(j-1)}_{\rm x \mid y}(\cdot \mid y), f\# q^{(j)}_{\rm x \mid y}(\cdot \mid y)) \\
        &< i \cdot \varepsilon.
    \end{align*}
    Thus together, we have for any $y \in \mathcal{Y}$
    \begin{align}
        W_{1, \ell}(f\# p_{\rm x \mid y}(\cdot \mid y), f\# q_{\rm x \mid y}(\cdot \mid y))
        &< \varepsilon \cdot \sum_{i=1}^s i \cdot r_{i} \nonumber \\
        &< \varepsilon \cdot a \cdot  \sum_{i=1}^s i \nonumber \\
        &< \varepsilon \cdot a \cdot \frac{s(s+1)}{2}.
        \label{eq:gs-w-bound-pred}
    \end{align}
    where the last inequality holds from taking an arithmetic sequence. Using the definition of the $\CCA(\kappa)$ assumption, the result follows immediately.
\end{proof}

\subsection{Proof of Theorem~\ref{thm:predentangle-upper}}

\begin{proof}
    To prove our upper bound on the label entanglement, we will utilize Theorem~\ref{thm:entangle-convert} and first upper bound the prediction entanglement. Consider the following decomposition of the prediction entanglement:
    \begin{align*}
        \predentangle(f) 
        &= \int W_{1, \ell}(p_{\rm f \mid y}(\cdot \mid y), q_{\rm f \mid y}(\cdot \mid y')) \dd \gamma_{\rm y}^\star(y, y') \\
        &= \int \llbracket y \neq y' \rrbracket \cdot W_{1, \ell}(p_{\rm f \mid y}(\cdot \mid y), q_{\rm f \mid y}(\cdot \mid y')) \dd \gamma_{\rm y}^\star(y, y')
        + \int \llbracket y = y' \rrbracket \cdot W_{1, \ell}(p_{\rm f \mid y}(\cdot \mid y), q_{\rm f \mid y}(\cdot \mid y')) \dd \gamma_{\rm y}^\star(y, y').
    \end{align*}
    We upper bound the first term corresponding to the case of $y \neq y'$. For any $y, y' \in \mathcal{Y}$, we have
    \begin{align*}
        W_{1, \ell}(p_{\rm f \mid y}(\cdot \mid y), q_{\rm f \mid y}(\cdot \mid y'))
        &= W_{1, \ell}(f \# p_{\rm x \mid y}(\cdot \mid y), f \# q_{\rm x \mid y}(\cdot \mid y')) \\
        &\leq W_{1, \ell}(f \# p_{\rm x \mid y}(\cdot \mid y), f \# p_{\rm x \mid y}(\cdot \mid y')) + W_{1, \ell}(f \# p_{\rm x \mid y}(\cdot \mid y'), f \# q_{\rm x \mid y} (\cdot \mid y')) \\
        &= W_{1, \ell}(f \# p_{\rm x \mid y}(\cdot \mid y'), f \# q_{\rm x \mid y} (\cdot \mid y')) + \int \ell(\hat{y}, \hat{y}') \dd \gamma^{p}_{\rm f}(\hat{y}, \hat{y}' \mid y, y'),
    \end{align*}
    where $\gamma^{p}_{\rm f}(\hat{y}, \hat{y}' \mid y, y')$ denotes the optimal transport plan between $p_{\rm f \mid y}(\cdot \mid y)$ and $p_{\rm f \mid y}(\cdot \mid y')$.
    This latter term can be further bounded,
    \begin{align*}
        \int \ell(\hat{y}, \hat{y}') \dd \gamma^{p}_{\rm f}(\hat{y}, \hat{y}' \mid y, y')
        &\leq \int \ell(\hat{y}, \hat{y}') \dd f\#\tilde{\gamma}^{p}_{\rm x}(\hat{y}, \hat{y}' \mid y, y') \\
        &= \int \ell(f(x), f(x')) \dd \tilde{\gamma}^{p}_{\rm x}(x, x' \mid y, y'),
    \end{align*}
    where the inequality follows from the surjectivity of the pushforward (similar reasoning as per proof of Lemma~\ref{lemma1}). The coupling $\tilde{\gamma}^{p}_{\rm x}$ corresponds to the resulting coupling of said pushforward.  

    
    %

    Now putting the above upper-bounds together, we can bound the prediction entanglement as follows:
    \begin{align*}
        \predentangle(f) 
        &\leq \int \llbracket y \neq y' \rrbracket \cdot \left( W_{1, \ell}(p_{\rm f \mid y}(\cdot \mid y'), q_{\rm f \mid y}(\cdot \mid y')) + L \right)  \dd \gamma_{\rm y}^\star(y, y') \\
        & \quad + \int \llbracket y = y' \rrbracket \cdot W_{1, \ell}(p_{\rm f \mid y}(\cdot \mid y'), q_{\rm f \mid y}(\cdot \mid y')) \dd \gamma_{\rm y}^\star(y, y') \\
        &= \int W_{1, \ell}(p_{\rm f \mid y}(\cdot \mid y'), q_{\rm f \mid y}(\cdot \mid y')) \dd \gamma_{\rm y}^\star(y, y') + L \cdot \int \llbracket y \neq y' \rrbracket  \dd \gamma_{\rm y}^\star(y, y'),
    \end{align*}
    where we used the assumed upper bound $L$ on the loss function. Now to finalize the upper bound on prediction entanglement, we make two additional observations.
    First, from \eqref{eq:gs-w-bound-pred} in Lemma~\ref{lem:gs-cca}'s proof, for any $y' \in \mathcal{Y}$, we have
    \begin{equation*}
        W_{1, \ell}(p_{\rm f \mid y}(\cdot \mid y'), p_{\rm f \mid y}(\cdot \mid y')) \leq \kappa' \defeq \varepsilon \cdot \frac{a}{2} s (s+1).
    \end{equation*}
    Second, from the lower bound on the loss $\ell$ for $y \neq y'$, we have,
    \begin{equation*}
        \int \llbracket y \neq y' \rrbracket \cdot \dd \gamma_{\rm y}^\star(y, y') \leq \frac{W_{1, \ell}(p_{\rm y}, q_{\rm y})}{l}.
    \end{equation*}
    This finally gives us,
    \begin{align*}
        \predentangle(f) 
        &\leq \int W_{1, \ell}(p_{\rm f \mid y}(\cdot \mid y'), q_{\rm f \mid y}(\cdot \mid y')) \dd \gamma_{\rm y}^\star(y, y') + L \cdot \int \llbracket y \neq y' \rrbracket  \dd \gamma_{\rm y}^\star(y, y') \\
        &\leq \kappa' + L \cdot \frac{\delta}{l}.
    \end{align*}
    Note that via assumption, we have $W_{1, \ell}(p_{\rm y}, q_{\rm y}) < \delta$. Now switching to label entanglement, from Theorem~\ref{thm:entangle-convert}, we have that
    \begin{align*}
        \labelentangle(f) 
        &\leq \predentangle(f) + W_{1, \ell}(p_{\rm y}, q_{\rm y}) + \risk_p(f) + \risk_q(f) \\
        &= \kappa' + \delta \cdot \frac{L + l}{l} + \risk_p(f) + \risk_q(f).
    \end{align*}
    
    Using the second bound in Corollary \ref{cor0}, we can bound the target risk $\risk_q(f)$ as
    \begin{equation*}
        \risk_q(f) \leq \risk_p(f) + W_{1,l}(p_y, q_y) + \predentangle(f) \leq b + \delta + \kappa' + L \cdot \frac{\delta}{l},
    \end{equation*}

    which gives us,
    \begin{align*}
        \labelentangle(f)
        \leq 2\kappa' + 2b + 2\delta \cdot \frac{L + l}{l}.
    \end{align*}
    Substituting $\kappa'$ yields the desired result.
\end{proof}

%% file: appendix/related_work.tex
\section{Related Work}\label{app:related_work}

Besides the UDA approaches mentioned in Section~\ref{related}, Domain Adaptation (DA) was also studied in a number of specialized contexts, ranging from semi-supervised DA (where we can expect a limited amount of target labels \citep{xu2019d}), and gradual DA (where the distributions typically shift slowly over time \citep{kumar2020understanding}), to test-time DA (where source domain inputs are not accessible at test-time \citep{nado2020evaluating}, \eg, due to privacy concerns) and universal DA (where the domain can enlarge to contain new classes) \citep{you2019universal}.

\subsection{Comparisons to Existing Theory}

When compared to previous work, we avoid removing the role of the loss $\ell$
and model $f$ from the statistical divergence term. Indeed, prior work
involving the Wasserstein distance has involved either partially removing the presence of
the model by assuming that a probabilistic Lipschitz property holds for
$f$~\citep[Theorem 3.1]{courty2017}; or assumes that the mapping $(x, y)
\mapsto \ell(f(x), y)$ is Lipschitz itself~\citep[Theorem 4.4]{wang2022information}. For the latter case of \citet{wang2022information}, Assumption~\ref{assum:metric}
is not needed to establish a similar bound to Lemma~\ref{lemma1}; however, Assumption~\ref{assum:metric} is eventually made for further theoretic
analysis. 

From a technical perspective, all these bounds involving the Wasserstein
distance (ours included) are based on exploiting the Lipschitz constant of
different metric spaces. Indeed, in our case, the cost function is already a
metric and thus is 1-Lipschitz to itself. See Appendix \ref{app:approx_triangle_ineq} for how to relax Assumption \ref{assum:metric} while maintaining the essential properties of the proved bounds. 

\subsection{Relationship to Other Change of Measure Inequalities}

A key point of comparison between our work and other domain matching methods which utilize divergences (over marginals) is the change of measure inequality used to switch from an expectation in $q$ to an expectation in $p$ (\ie, switching from the target risk to the source risk via an upper bound). In general, the change of measure inequalities yield variants of Lemma~\ref{lemma1} that use different divergences (dissimilarity functions).

Our work is based on Wasserstein distances that use a particular choice of (decomposable) cost function derived by the loss function $\ell$ of the prediction task, Lemma~\ref{lemma1}. Several other approaches also utilize the Wasserstein distance, but differ in the specific cost functions used~\citep{courty2017,damodaran2018deepjdot,wang2022information}. One can, as a corollary, derive many of these variants of Lemma~\ref{lemma1} by making additional Lipschitz continuity assumptions for the loss function $\ell$ or the model $f$.

Alternatively, a common choice of dissimilarity function is the KL-divergence via the Donskar-Varadhan change of measure inequality~\citep{donsker1983asymptotic}. With the addition of the sub-Gaussian loss assumption, square-root of the KL-divergence was used in domain adaptation~\citep{nguyen2022kl,wang2022information}. More recent work has also modified the variational form of f-divergences to achieve similar upper bounds~\citep{acuna2021f,wang2024f}. We discuss the relation of our Wasserstein distance based bounds to the KL-divergence in more detail in Appendix \ref{add_remarks}.

\subsection{Relationship to Other Conditional Discrepancy Terms}

The entanglement terms that we proposed in section \ref{intro_entanglement} are examples of \emph{conditional discrepancy} terms introduced in the literature. 
An \emph{average} conditional discrepancy term was first introduced in Theorem 1 of \citet{benDavid2010}, however it was not used in the central theorem of that paper (Theorem 2) that utilized the $H \Delta H$ divergence to bound the target risk. The authors in \citet{zhao2019learning} proposed (Theorem 4.1) using the conditional discrepancy term in the proof of Theorem 2 of \citet{benDavid2010}, thereby replacing the $\lambda$ of $\LJE(\lambda)$. This quantity, the joint error of the best performing hypothesis in the hypothesis class, was shown by \citet{johansson2019support} to be not invariant over the (typically) non-invertible layers of deep neural networks. 

The Error Decomposition Theorem 3.1 of \citet{combes2020} forms a counterpart to the Theorem 4.1 in \citet{zhao2019learning}, in that the latter corresponds to the first bound \eqref{eq:bound_in_x} of Corollary \ref{cor0} in our paper, while the former corresponds to the second bound \eqref{eq:bound_in_y}. While we used the bound \eqref{eq:bound_in_y} only for theoretical analysis in Section \ref{sec:entanglement}, the authors in \citet{combes2020} use Theorem 3.1 to come up with a weighted domain matching algorithm. This approach was justified by making a \emph{Generalized Label Shift} (GLS) assumption \citep{combes2020}, which is a sufficient condition to enforce zero (prediction) entanglement, however, it is in general not possible to satisfy such an assumption without target labels (as can be seen by the dependency of the upper bound in Theorem 3.4 of \citep{combes2020} on the target risk). The entanglement terms thus capture the hardness of UDA (from an information-theoretic perspective) in a concise but general way: during any domain matching procedure that does not have access to target labels, the optimized model
may entangle the conditionals (in the output space), and we cannot expect any assumptions such as GLS to be a useful guide during this optimization. 

The authors in \cite{federici2021an} use the KL-divergence chain rule and mutual information to introduce their \emph{latent concept shift}, which can be viewed as another conditional discrepancy term. The introduced bounds utilize the KL-divergence and its symmetric counterpart the JS-divergence, both of which require the shared support assumption, unlike the optimal-transport based bounds. We highlight the connection between optimal transport and KL-divergence based bounds in more detail in Appendix~\ref{app:related_work}.

As for works that utilized Wasserstein distances, \citet{courty2017, redko2017, shen2018wasserstein, damodaran2018deepjdot} all used the $\lambda$ of the low joint error assumption ($\LJE(\lambda$) in place of the entanglement terms appearing in our bounds. Our analysis, derived through the simple bound of Lemma \ref{lemma1} and the decomposition rules in \ref{lemma2}, was not considered before in its generality. 

Similar to \citet{combes2020}, \citet{kirchmeyer2022mapping} also proposes an IW-based framework to handle both label shift \emph{and} conditional shift. Yet unlike most UDA methods, \citet{kirchmeyer2022mapping} does not constrain the model to learn only invariant representations  but trains a target classifier directly on transported source inputs. See also \citet{Shu2018ADA} for an approach which considers adapting a domain invariant model on unlabeled target inputs (using pseudolabeling) assuming potential violations of the $\LJE(\lambda)$ assumption.




%% file: appendix/approximate-triangle-inequality.tex
\newcommand{\approxtri}{\varkappa}

\section{Approximate Triangle Inequality}\label{app:approx_triangle_ineq}

In this section we study how we can relax the Assumption~\ref{assum:metric} while maintaining the essential properties of the proved bounds. 
A key property of Assumption~\ref{assum:metric} is the triangle inequality axiom. This is crucial for proving Lemma~\ref{lemma1} without removing the occurrence of $\ell$. In the latter case, removing $\ell$ is typically done via the Lipschitz continuity of $\ell$: but this simply switches the notion of distance (originally defined by a semi-metric $\ell$) to the underlying metric space of $\mathcal{X} \times \mathcal{Y}$ (which does have a triangle inequality property).

Many loss function violate the simple triangle inequality, but can satisfy a weaker condition.
To weaken the metric assumption, one can consider an approximate triangle inequality.

\begin{definition}[{$ \approxtri $-Approximate Triangle Inequality~\citep{crammer2008learning}}]
    \label{def:approx-triangle}
    Let $d \colon \mathcal{Z} \times \mathcal{Z} \to \mathbb{R}$ be a function. It satisfies the $\approxtri$-Approximate Triangle Inequality for $\approxtri \geq 1$ if,
    \begin{equation}
        d(z, z') \leq \approxtri \cdot \left( d(z, z'') + d(z'', z') \right),
    \end{equation}
    for all $z, z', z'' \in \mathcal{Z}$.
\end{definition}

Note that $\approxtri = 1$ reduces Definition~\ref{def:approx-triangle} to the usual triangle inequality.

In prior work, \citet{wang2022information} explored the approximate triangle inequality in the context of domain adaptation. However, this was specifically for dealing with the $\LJE(\lambda)$ assumption.

We note that many of our results hold when utilizing the following weaker assumption.

\begin{assumption}[$\approxtri$-Almost-Metric Loss]
    \label{assum:almost-metric}
    The loss function $\ell$ is a semi-metric (i.e., Assumption~\ref{assum:metric} without the triangle inequality) and also satisfies Definition~\ref{def:approx-triangle} with $\approxtri$.
\end{assumption}

This immediately helps us to obtain a variant of our key Lemma~\ref{lemma1}.

\begin{lemma}\label{lem:com-almost-metric}
    Suppose the loss function $\ell \colon \mathcal{Y} \times \mathcal{Y}
    \rightarrow \mathbb{R}$ satisfies Assumption~\ref{assum:almost-metric} with $\approxtri \geq 1$. Then the target \emph{risk} of a
    classifier $f \colon \mathcal{X} \rightarrow \mathcal{Y}$ is bounded by:
    \begin{align}
        \risk_{q}(f) 
        &\leq \approxtri^2 \cdot \risk_{p}(f) + \approxtri \cdot W_{1, \ell \circ f, \approxtri \ell}(p, q) \\
        &\overset{\rm (s)}{\leq} \approxtri^2 \cdot \risk_{p}(f) + \approxtri \cdot W_{1, \ell, \approxtri \ell}(f\#p, f\#q),
    \end{align}
    where the inequality $\rm (s)$ holds for surjective $f$. Additionally, $\rm (s)$ is an \emph{equality} whenever $f$ is invertible.
    We remind that 
    $(f\#p)(\hat{y}, y) = p( f^{-1}(\hat{y}), y)$.
\end{lemma}
\begin{proof}[Proof of Lemma~\ref{lem:com-almost-metric}]
    The proof follows identically to Lemma~\ref{lemma1}, except for the initial application of the triangle inequality. Examine the following for $\gamma \in \Gamma(q, p)$,
    \begin{align*}
        \risk_{q}(f) &= \expect_{q}[\ell(f(x),y)] \\
        &= \approxtri^2 \cdot \expect_{p}[\ell(f(x),y)] + \expect_{(x, y), (x', y') \sim \gamma}\left[ \ell(f(x'),y') - \approxtri^2 \cdot \ell(f(x),y) \right] \\
        &\leq \approxtri^2 \cdot \expect_{p}[\ell(f(x),y)] + \expect_{(x, y), (x', y') \sim \gamma}\left[ \approxtri \cdot \ell(f(x),f(x')) + \approxtri \cdot \ell(f(x), y') - \approxtri^2 \cdot \ell(f(x),y) \right] \\
        &\leq \approxtri^2 \cdot \expect_{p}[\ell(f(x),y)] + \expect_{(x, y), (x', y') \sim \gamma}\left[ \approxtri \cdot \ell(f(x),f(x')) + \approxtri^2 \cdot \ell(f(x), y) + \approxtri^2 \cdot \ell(y, y') - \approxtri^2 \cdot \ell(f(x),y) \right] \\
        &= \approxtri^2 \cdot \expect_{p}[\ell(f(x),y)] + \expect_{(x, y), (x', y') \sim \gamma}\left[ \approxtri \cdot \ell(f(x),f(x')) + \approxtri^2 \cdot \ell(y, y')  \right] \\
        &= \approxtri^2 \cdot \expect_{p}[\ell(f(x),y)] + \approxtri \cdot \expect_{(x, y), (x', y') \sim \gamma}\left[ \ell(f(x),f(x')) + \approxtri \cdot \ell(y, y')  \right].
    \end{align*}
    Then, with this change, the proof of Lemma~\ref{lemma1} follows without change to yield the result.
\end{proof}

To further see how switching from Assumption~\ref{assum:metric} to Assumption~\ref{assum:almost-metric} affects our results,  we also derive an equivalence to Corollary~\ref{cor0}.

\begin{corollary}\label{cor:com-entanglements-almost-metric}
    Suppose the loss function $\ell \colon \mathcal{Y} \times \mathcal{Y}
    \rightarrow \mathbb{R}$ satisfies Assumption~\ref{assum:almost-metric} with $\approxtri \geq 1$ and a classifier $f \colon \mathcal{X} \rightarrow \mathcal{Y}$ satisfies Assumption~\ref{assum:surjective} then
    \begin{align}
        \risk_{q}(f) &\leq \approxtri^2 \cdot \risk_{p}(f) + \approxtri \cdot W_{1, \ell}(f\#p_{\rm x}, f\#q_{\rm x}) 
        +
        \approxtri^2 \cdot \labelentangle(f),
        \\
        \risk_{q}(f) &\leq \approxtri^2 \cdot \risk_{p}(f) + \approxtri^2 \cdot W_{1, \ell}(p_{\rm y}, q_{\rm y}) 
        +
        \approxtri \cdot 
        \predentangle(f).
    \end{align}
\end{corollary}
\begin{proof}[Proof of Corollary~\ref{cor:com-entanglements-almost-metric}]
    Result follows immediately from Lemma~\ref{lem:com-almost-metric} and Lemma~\ref{lemma2} whilst noticing that $W_{1, \approxtri \ell} = \approxtri \cdot W_{1, \ell}$.
\end{proof}

Our remaining results follow similarly. Note that a consequence of the difference between Corollary~\ref{cor0} and Corollary~\ref{cor:com-entanglements-almost-metric} is that the Oracle Upper Bound (OUB) $\mathcal{U}$ changes to
\begin{equation}
    \mathcal{U}_{\approxtri}(f) \defeq \approxtri^2 \cdot \risk_{p}(f) + \approxtri \cdot W_{1, \ell}(f\#p_{\rm x}, f\#q_{\rm x}) 
    +
    \approxtri^2 \cdot \labelentangle(f).
\end{equation}
Notice that as $\approxtri$ increases, so does the influence of the label-entanglement term with respect to the marginal distance. Also from an optimization perspective, the Wasserstein Regularized Risk (WRR) equivalent
\begin{equation}
    \mathcal{R}_{\approxtri}(f) \defeq \approxtri^2 \cdot \risk_{p}(f) + \approxtri \cdot W_{1, \ell}(f\#p_{\rm x}, f\#q_{\rm x}),
\end{equation}
now requires a scaling by $\approxtri$ depending on the $\approxtri$-approximate triangle inequality. This scaling is equivalent to setting the correct regularization constant when optimizing WRR.

%% file: appendix/remarks.tex
\section{Additional Remarks}\label{add_remarks}

\paragraph{Remark.} An interesting case of the bounds in Lemma~\ref{lemma2} occurs when the marginal source and target distributions are identical. For example, if $p_{\rm y} =
q_{\rm y}$, then the first term of Eq.~\eqref{eq:lem-decomp-y} vanishes and
$\gamma^\star_{\rm y}(y,y')$ is diagonal. As a result, the inequality further reduces to $W_{\alpha, c}(p, q) \leq \expect_{y \sim p_{\rm y}}\big[
W^\alpha_{\alpha, c_2}(p_{\rm x \mid y}(\cdot \mid y), q_{\rm x \mid y}(\cdot
\mid y))\big]^{1/\alpha}$. Moreover, the results can be upper bounded further using
any feasible coupling, including, e.g., the Cartesian coupling $(x, x') \mapsto p(x)q(x')$ as opposed to $\gamma_x^{\star}$ in \eqref{eq:lem-decomp-x}. 


\paragraph{Close Conditionals Assumption.}

We give an explicit construction to show that CCA is a valid assumption for networks with large enough capacity. Assume that there's a network $f \colon \mathcal{X} \to \mathcal{Y} \subseteq \mathbb{R}^{n}$ that satisfies the LJE assumption, i.e., $\lambda = \min_f \risk_p(f) + \risk_q(f) $ is low for such a network. We will compose the output of the network with a nonlinear mapping $g(\cdot)$ such that the resulting enlarged network $h = g \circ f$ satisfies CCA: $\kappa = \min_h \risk_p(h) + \max_y W_{1,\ell}(h \# p(x|y), h \# q(x|y))$ for some low $\kappa$. 

Consider the function $g(\cdot)$ composed of mappings $g_i(\cdot)$ such that $g = g_{m} \circ g_{m-1} \ldots \circ g_{1}$, where for each $g_i = g_{in} \circ g_{i(n-1)} \ldots \circ g_{i1}$, $i \in \mathbb{Z}^{m}$, $g_{ij}$ switches between affine operations
\begin{align*}
    g_{ij}(x) = \begin{cases}
        A_{ij} x + b_{ij} & \text{if } x_j > x_k \text{ for all } k \neq j, \\
        x & \text{else.}
    \end{cases}
\end{align*}
Such a mapping can be implemented approximately using, e.g., a ReLU network that uses an \emph{and} logic internally. The affine operations $A_{ij} x + b_{ij}$ can be defined such that each decreases the Wasserstein distance between the $j$-th conditionals sufficiently, while making sure that the conditionals are still linearly separable from the rest, e.g., a mapping that shrinks the two conditionals closer towards their mutual mean.

\paragraph{Relation to KL-divergence.} It is possible to relate Lemma~\ref{lemma1} to the KL-divergence by: (1) upper-bounding $W_{1, \ell, \ell}$ via a Lipschitz continuity assumption on $\ell$ to get $W_{1, d}$, where $d$ is the metric distance of $\mathcal{X} \times \mathcal{Y}$; (2) further bounding the Wasserstein distance with the total variation (TV) distance, taking into account the diameter of the space; and finally (3) using the Pinsker's inequality~\citep{pinsker1964information} or the Bretagnolle-Huber bound~\citep{bretagnolle1979estimation}. For the last step, one can take the minimum of both inequalities as they do not strictly dominate each other~\citep{canonne2022short}. A specific combination of these steps has been previously leveraged in~\citet{wang2022information}.

In what follows, we present a version of Corollary~\ref{cor0} which switches from our Wasserstein distance to a KL-divergence. Instead of doing both (1) and (2) in the above, we consider the diameter of the space \wrt the specific metric used in the Wasserstein distance $W_{1, \ell}$.
\begin{corollary}
    Suppose that the pre-conditions of Corollary~\ref{cor0} hold and that $\ell$ satisfies Assumption~\ref{assum:bounded}. Then
    \begin{align}
        \risk_{q}(f) &\leq \risk_{p}(f) + L \cdot \sqrt{\frac{1}{2} \cdot \min \left\{ \KL(f\#p_{\rm x} \Mid f\#q_{\rm x}), \KL(f\#q_{\rm x} \Mid f\#p_{\rm x}) \right\} }
        +
        \labelentangle(f)
        \\
        \risk_{q}(f) &\leq \risk_{p}(f) + L \cdot \sqrt{\frac{1}{2} \cdot \min\left\{ 
        \KL(p_{\rm y} \Mid q_{\rm y}),
        \KL(q_{\rm y} \Mid p_{\rm y})
        \right\}
        }
        +
        \predentangle(f).
    \end{align}
\end{corollary}
\begin{proof}
    Follows immediately from bounding the diameter of $\mathcal{Y}$ \wrt the bounded loss $\ell$ and Pinsker's inequality.
\end{proof}

One could generalize the above by replacing $L$ with the diameter of $\mathcal{Y}$ under the metric $\ell$. A practical consequence of the above corollary is that one could potentially switch the Wasserstein regularized risk (WRR) to use the above upper bound, i.e., modifying WRR to
\begin{equation}
    \mathcal{R}_\KL(f) \defeq
    \risk_p(f) + L \cdot \sqrt{\frac{1}{2} \cdot  
    \min \left\{ 
        \KL(f\#p_{\rm x} \Mid f\#q_{\rm x}),
        \KL(f\#q_{\rm x} \Mid f\#p_{\rm x}) 
    \right\}
    }
    \leq 
    \risk_p(f) + L \cdot \sqrt{\frac{1}{2} \cdot \JS(f\#p_{\rm x} \Mid f\#q_{\rm x})},
\end{equation}
presents an alternative divergence to minimize in practice. One could also extend the above corollary to define a KL-divergence version of the entanglement term. One should, however, note that KL-divergences evaluate to $\infty$ whenever the support of the distributions considered are disjoint. 

\paragraph{Gaussians.}

We explore the bound \eqref{eq:lem-decomp-x} in Lemma~\ref{lemma2} in the case of Gaussians with scaled covariances, using the Euclidean distance for $\ell$ and setting $\alpha = 2$. The arguments apply due to symmetry (by switching $y \leftrightarrow x$) to \eqref{eq:lem-decomp-y} as well.

Assume that $x, y \sim p = \mathcal{N}(\mu, \Sigma)$ and $x', y' \sim q = \mathcal{N}(\mu', \Sigma')$ where
\begin{align*}
    \mu &\defeq (\mu_x^{\mathrm{T}}, \mu_y^{\mathrm{T}})^{\mathrm{T}}, \\
    \Sigma &\defeq \begin{pmatrix}
        \Sigma_x & \Sigma_{xy} \\
        \Sigma_{yx} & \Sigma_y
    \end{pmatrix},
\end{align*}
with corresponding notation for $\mu', \Sigma'$. We first start by showing that whenever $\Sigma' = s^{2} \Sigma$ for some $s \in \mathbb{R}$, the Wasserstein-2 distances satisfy
\begin{align*}
    W_{2,\ell}^{2}(p, q) = W_{2,\ell}^2(p_y, q_y) + \int \pi^{\star}(y, y') W_{2,\ell}^{2}(p_{x \mid y}(\cdot \mid y), q_{x \mid y}(\cdot \mid y')) \dd (y, y'),
\end{align*}
where $p(x,y)$ is decomposed as $p(x,y) = p_y(y)p_{x|y}(x|y)$ (sym. $q$) and $\pi^{\star}(y, y')$ is the optimal transport map between $p_y, q_y$. The Wasserstein-2 distance between the joint Gaussians $p, q$ is given by \citep{peyre2019computational}
\begin{align*}
    W_{2,\ell}^{2}(p, q) = \|\mu - \mu'\|^{2} + \tr(\Sigma + \Sigma' - 2(\Sigma^{1/2}\Sigma'\Sigma^{1/2})^{1/2}).
\end{align*}
Plugging in $\Sigma' = s^{2}\Sigma$ we get
\begin{align*}
    W_{2,\ell}^{2}(p, q) &= \|\mu - \mu'\|^{2} + (s - 1)^{2}\tr(\Sigma), \\
    W_{2,\ell}^{2}(p_y, q_y) &= \|\mu_y - \mu_y'\|^{2} + (s - 1)^{2}\tr(\Sigma_y),
\end{align*}
where the distance between the marginals $p_y, q_y$ has the optimal transport given by a deterministic mapping $T(y)$
\begin{align*}
    \pi^{\star}(y, y') &= \delta_{(y, T(y))}, \\
    T(y) &\defeq \mu_y' + \Sigma_y^{-1/2}\left(\Sigma_y^{1/2}\Sigma_y'\Sigma_y^{1/2}\right)^{1/2}\Sigma_y^{-1/2}(y - \mu_y) \\
    &= \mu_y' + s(y - \mu_y).
\end{align*}
Given that $\mu_{x \mid y} = \mu_x + \Sigma_{xy}\Sigma_y^{-1}(y - \mu_y)$ and $\Sigma_{x \mid y} = \Sigma_x - \Sigma_{xy} \Sigma_y^{-1}\Sigma_{yx}$, the 2-Wasserstein distance between $p(x \mid y)$ and $q(x \mid y)$ is given by
\begin{align*}
    W_{2,\ell}^{2}(p_{x \mid y}, q_{x \mid y}) = \| \left(\mu_x + \Sigma_{xy}\Sigma_y^{-1}(y - \mu_y)\right) - \left(\mu_x' + \Sigma_{xy}\Sigma_y^{-1}(y' - \mu_y')\right)\|^{2} + (s - 1)^{2}\tr(\Sigma_{x \mid y}).
\end{align*}
Plugging into the integral above and using $T(y)$ we get
\begin{align*}
&\int W_{2,\ell}^{2}(p_{x \mid y}(\cdot \mid y), q_{x \mid y}(\cdot \mid \mu_y' + s(y - \mu_y))) \dd y \\
= &\int \|\left(\mu_x + \Sigma_{xy}\Sigma_y^{-1}(y - \mu_y)\right) - \left(\mu_x' + \Sigma_{xy}\Sigma_y^{-1}(\mu_y' + s(y - \mu_y) - \mu_y'\right)\|^{2} \dd y + (s-1)^{2}\tr(\Sigma_{x \mid y}) \\
= & \, \|\mu_x - \mu_x' \|^{2} + 2(1-s)(\mu_x - \mu_x')^{\mathrm{T}}\Sigma_{xy}\Sigma_y^{-1} \int (y - \mu_y) \dd y + \\
& (s-1)^{2} \tr\left((\Sigma_{xy}\Sigma_y^{-1})^{\mathrm{T}}(\Sigma_{xy}\Sigma_y^{-1})\int (y - \mu_y)(y - \mu_y)^{\mathrm{T}} \dd y \right) + (s-1)^{2}\tr(\Sigma_{x \mid y}) \\
= & \|\mu_x - \mu_x' \|^{2} + (s-1)^{2}\tr(\Sigma_y^{-1}\Sigma_{yx}\Sigma_{xy} + \Sigma_{x \mid y}),
\end{align*}
since $\int (y - \mu_y) \dd y = 0$ and $\int (y - \mu_y)(y - \mu_y)^{\mathrm{T}} \dd y = \Sigma_y$. Now since $\tr(ABC) = \tr(BCA)$ for any matrix triple $A, B, C$, $\tr(\Sigma_{x \mid y})$ can be written as $ \tr(\Sigma_{x \mid y}) = \tr(\Sigma_x) - \tr(\Sigma_y^{-1}\Sigma_{yx}\Sigma_{xy})$. Plugging into the above and adding $W_2^{2}(p_y, q_y)$ we get
\begin{align*}
    W_{2,\ell}^2(p_y, q_y) + & \int \pi^{*}(y, y') W_{2,\ell}^{2}(p_{x \mid y}(\cdot \mid y), q_{x \mid y}(\cdot \mid y')) \dd(y, y') \\
    & = \|\mu_y - \mu_y' \|^{2} + (s-1)^{2}\tr(\Sigma_y) + \|\mu_x - \mu_x' \|^{2} + (s-1)^{2}\tr(\Sigma_x) \\
    & = \|\mu - \mu' \|^{2} + (s-1)^{2}\tr(\Sigma) \\
    & = W_{2,\ell}^{2}(p, q),
\end{align*}
which proves the above. Now we relate (by switching between the couplings) the joint Wasserstein distance in Lemma \ref{lemma2} that uses the composite cost $c(x, y, x', y') = \ell(x, x') + \ell(y, y')$ to $W_{2,\ell}^{2}(p, q)$ 
\begin{align*}
    W_{2,\ell,\ell}^{2}(p, q) \leq W_{2,\ell}^{2}(p, q) + 2 \int \ell(x, x')\ell(y, y') \pi^{\star}(x,y,x',y') \dd((x, y), (x', y')),
\end{align*}
where $\pi^{\star}(x,y,x',y')$ is the optimal map corresponding to $W_{2,\ell}(p,q)$ and is decomposed (due to $\Sigma' = s^{2}\Sigma$) as
\begin{align*}
    &\pi^{\star}(x,y,x',y') = \delta_{x, T_1(x)} \delta_{y, T_2(y)}, \\
    &T_1(x) = \mu_x' + s(x - \mu_x), \\
    &T_2(y) = \mu_y' + s(y - \mu_y).
\end{align*}
Hence the integral above can be written as
\begin{align*}
    \int \ell(x,x')\ell(y,y') \pi^{\star}(x, y, x', y') \dd((x, y), (x', y')) = \int \ell(x, T_1(x))\ell(y, T_2(y)) \dd(x,y).
\end{align*}
Applying Cauchy-Schwartz inequality we see that finally
\begin{align*}
    W^{2}_{2,\ell,\ell}(p,q) \leq W_{2,\ell}^{2}(p_y, q_y) + \expect_{\pi^{\star}(y, y')}[W_2^{2}(p_{x|y}(\cdot \mid y), q_{x|y}(\cdot \mid y'))] + 2 W_{2,\ell}(p_y, q_y)\left(\expect_{\pi^{\star}(y, y')}[W_2^{2}(p_{x|y}(\cdot \mid y), q_{x|y}(\cdot \mid y'))]\right)^{1/2}.
\end{align*}
Taking square root of both sides we get the inequality \eqref{eq:lem-decomp-x} for $\alpha = 2$.

%% file: appendix/experiment_details.tex
\section{Experimental Details}\label{sec:experiment_details}

In this section we discuss in more detail the experimental setup used to produce the results in Section~\ref{sec:experiments}. We generate accuracy plots over the epochs for the methods shown in Table \ref{table1}, together with entanglement estimates, to support the conclusions that were drawn.

\paragraph{Scenarios.} The scenarios considered are: MNIST to USPS, USPS to MNIST, MNIST to MNIST-M, SVHN to MNIST, CIFAR-10 corruptions and the Portraits dataset. MNIST and USPS are standard \emph{digits} datasets, with handwritten digit styles showing significant variations between them. SVHN is a more challenging dataset composed of colored street view images of house numbers. MNIST-M is a dataset that is causally related to MNIST: it is created by changing the background of MNIST inputs with colored images from another dataset, see \citet{ganin2016} for details. Portraits is a set of yearbook images from early 1900s to early 2000s, in particular, it is sometimes presented as a dataset suitable for \emph{gradual} domain adaptation~\citep{kumar2020understanding}.\footnote{However preliminary experiments conducted using different methods, including pseudolabeling typically used in gradual domain adaptation, see no benefit in adapting to images gradually according to year, compared to domain matching that ignores the gradual shift.} In our case the source dataset uses labeled images of girls and boys until the year $1974$ (first half of the dataset) and the remaining half of the dataset (from $1974$ till $2013$) is used as the target. Hairstyles and lighting conditions, in particular, undergo significant change over the years. Since the Portraits dataset has only two classes, it is a convenient dataset to use for visualizing entanglement, see Figure~\ref{fig:portraits_entanglement}. Finally CIFAR-10 corruptions (CIFAR-10c) are created using different noise models, such as changes in illumination or motion blur~\citep{hendrycks2019benchmarking}. In our case we report the average result of the three most difficult (in terms of degradation of ERM accuracy) noise models, which are fog, frost and snow-based corruption models. 

For each scenario, we utilize the default train-test splits for source and target datasets if provided; otherwise we split the data using \%80/20 split ratio. When the input dimensions of the datasets do not match in a distribution shift scenario, we either downsample (e.g., in the case of MNIST $\to$ USPS, we resize the MNIST images to $16 \times 16$) or upsample (e.g., in the case of SVHN $\to$ MNIST, we resize the grayscale MNIST images to $32 \times 32 \times 3$). For the Portraits dataset, we resize the images down to $32 \times 32$\footnote{We also tried domain adaptation with the original sizes, however, the source and target accuracies were significantly lower in that case for the models that we use in this paper.}.

\paragraph{Models.} We run each UDA scenario on different models of varying complexity, in order to get a better picture of the dependency of entanglement on the model capacity and type. The first model that we test is a two hidden layer multilayer perceptron (MLP) with hidden layer neurons of size $(200, 100)$, referred to as \emph{MLP}. The rest are convolutional neural networks (CNN) with various depth sizes and regularizations: \emph{small\_cnn} is a CNN with two convolutional layers (with ReLU nonlinearity and max pooling) followed by two fully connected layers, \emph{LeNet} is a CNN with two convolutional layers using average pooling followed by three fully connected layers. The last two bigger networks used are referred to as \emph{conv1} and \emph{conv2}: \emph{conv1} is a deeper CNN than the others (six convolutional layers followed by two fully-connected layers) and \emph{conv2} (two convolutional layers followed by three fully-connected layers) has additional batch-norm and dropout regularization between the layers. The model \emph{conv2} was used to visualize the effect of entanglement during UDA in Figure~\ref{fig:portraits_entanglement}.

Due to space constraints, in Table~\ref{table1}, we present only the final source and target accuracies at the last epoch from the best model (\emph{conv1}) and choose another randomly (different for each scenario) from the models mentioned above that satisfy $\LJE$ oracle accuracy above $90 \% $ for both the source and target domains (except for the CIFAR-10 corruptions scenario). Fixing the final epoch can be misleading as an indicator of performance, as the accuracy curves tend to show significant oscillatory behaviour, which gives useful indication about the quality of the marginal alignment and the effect of the entanglement. Hence, in addition to the results presented in Table~\ref{table1}, in this section we plot the resulting accuracy curves over the epochs as well.

\paragraph{Methods.} There are five methods presented in Table~\ref{table1}. The $\LJE$ \emph{Oracle} trains the model on both source \emph{and} target labels and is provided here as a reference to check whether the $\LJE$ assumption holds to some degree. \emph{ERM} trains the model only on source labels, and similarly is provided as a reference to check for \emph{negative transfer} (i.e., the potential loss in accuracy after adapting to the target input distribution). \emph{DANN}~\citep{ganin2016} is an adversarial UDA approach that uses an additional \emph{discriminator} network (a two-layer CNN with batch-norm) to minimize the divergence between the source and target. DANN and related adversarial methods can be shown to approximately minimize a Jensen-Shannon divergence \citep{goodfellow2014gan}. \emph{WRR} minimizes the regularized source risk presented in \ref{def:w-regularized-risk}, without any adjustments such as scaling. \emph{JDOT}~\citep{courty2017, damodaran2018deepjdot} is a method that adds a  feature space Wasserstein distance minimization (with multiplier $\alpha = 0.001$) in addition WRR, which potentially trades off entanglement for a higher marginal distance. As we see in the experiments, this strategy to constrain the network Lipschitz bound and hence the entanglement, is sometimes effective (when compared to the vanilla WRR) but not always. WRR and JDOT both minimize cross-entropy as the loss function in Table~\ref{table1} but JDOT scales it by a factor of $0.001$.



In Table \ref{table2} we present an additional oracle that trains the model by optimizing the $\CCA$ objective function: $\min_f \risk_{p}(f) + \max_{y \in \mathcal{Y}} W_{1, \ell}(f\# p_{\rm x \mid \rm y}(\cdot \mid y), f\# q_{\rm x \mid \rm y}(\cdot \mid y))$. Although the $\max$ operator is not differentiable, Pytorch keeps track of the index of the conditionals that obtain the maximum distance and differentiates through it, which corresponds to a subdifferential. 

\paragraph{Libraries.} We use the Pytorch library~\citep{paszke2019} to optimize the Deep Neural Network models and for the implemention of the methods mentioned above. For the OT computations in WRR and JDOT, we use the GeomLoss library~\citep{feydy2019interpolating} which computes the entropy-regularized Wasserstein distance (\ie, the Sinkhorn divergence~\citep{peyre2019computational}), where we set the regularization to the default value of $0.05$. We use the 2-Wasserstein distance rather than the 1-Wasserstein distance for computational reasons, although the difference is generally small. To estimate the entanglement term in Table~\ref{table2} we use the POT library \citep{flamary2021pot}, as we need the optimal transport map $\gamma^{\star}(x, x')$ available as an output (GeomLoss makes available only the value and not the mapping).

\paragraph{Plots.} In Figures \ref{fig:acc_plots_table1_row1} to \ref{fig:acc_plots_table1_row3} we pick three scenarios and models to present target accuracy errorbar plots for each method shown in Table \ref{table1}. We present mean +/- standard deviations over three different seeds, over 10 epochs and plot on the right hand side the entanglement estimates of each method. The entanglement estimates are averaged over the mini-batches. It can be seen that the entanglement estimates have a strong negative correlation with the method's target accuracies. The ordering of the accuracies is also generally reversed in the case of entanglement.

In Figures \ref{fig:acc_plots_table2_row1} to \ref{fig:acc_plots_table2_row3} we pick three scenarios and models to plot the WRR performances using the Euclidean distance as the loss function. These scenario/model cases correspond to the results shown in Table \ref{table2}. The entanglement estimates together with the WRR values are plotted together with the target loss. The WRR values together with the entanglement estimates upper bound the target loss for all cases considered. Although we have not considered the optimization landscape of WRR in this article, results suggest that the WRR objective function is not minimized effectively. 

\include{figures/table1/acc_plots}
\include{figures/table2/acc_plots}


%% file: figures/table1/acc_plots.tex
\begin{figure}
    \centering
    \begin{minipage}{0.4\textwidth}
        \centering
        \includegraphics[width=\textwidth]{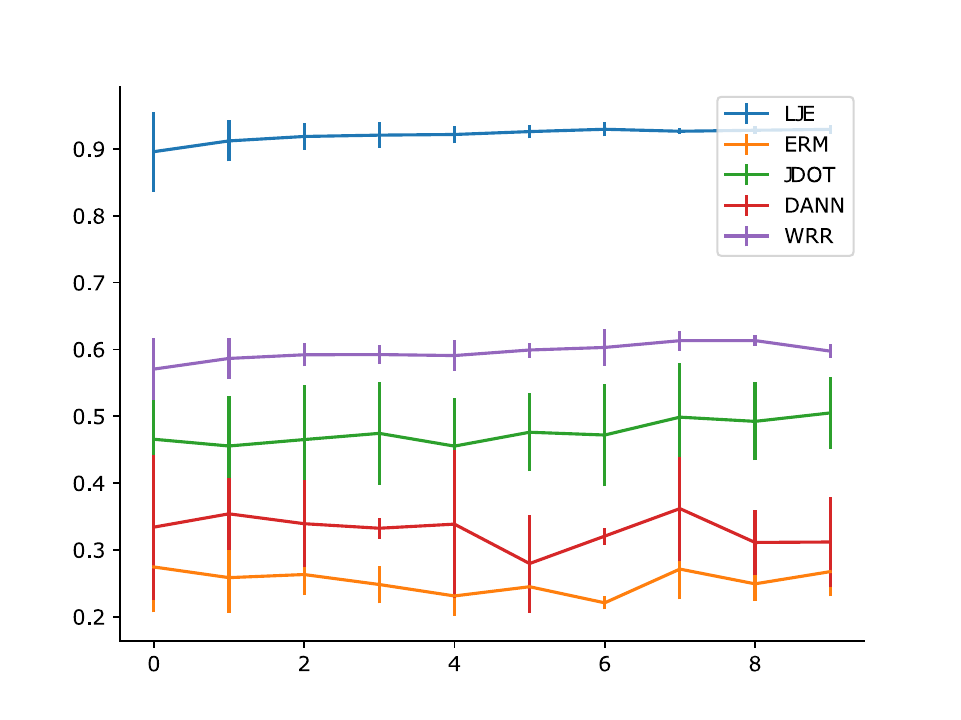}
    \end{minipage}
    \begin{minipage}{0.4\textwidth}
        \centering
        \includegraphics[width=\textwidth]{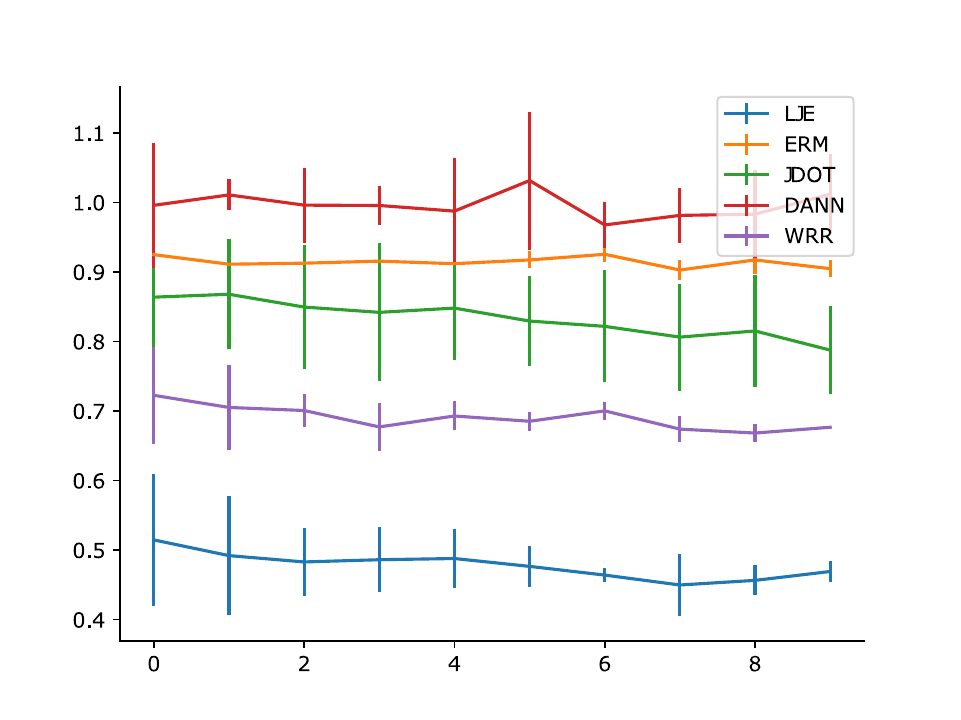}
    \end{minipage}
    \caption{Target accuracy plots for the MNIST $\to$ USPS scenario shown in Table \ref{table1} using the model MLP. Corresponding entanglement estimates are shown on the right hand figure.}
    \label{fig:acc_plots_table1_row1}
    \vspace{1em}
    \centering
    \begin{minipage}{0.4\textwidth}
        \centering
        \includegraphics[width=\textwidth]{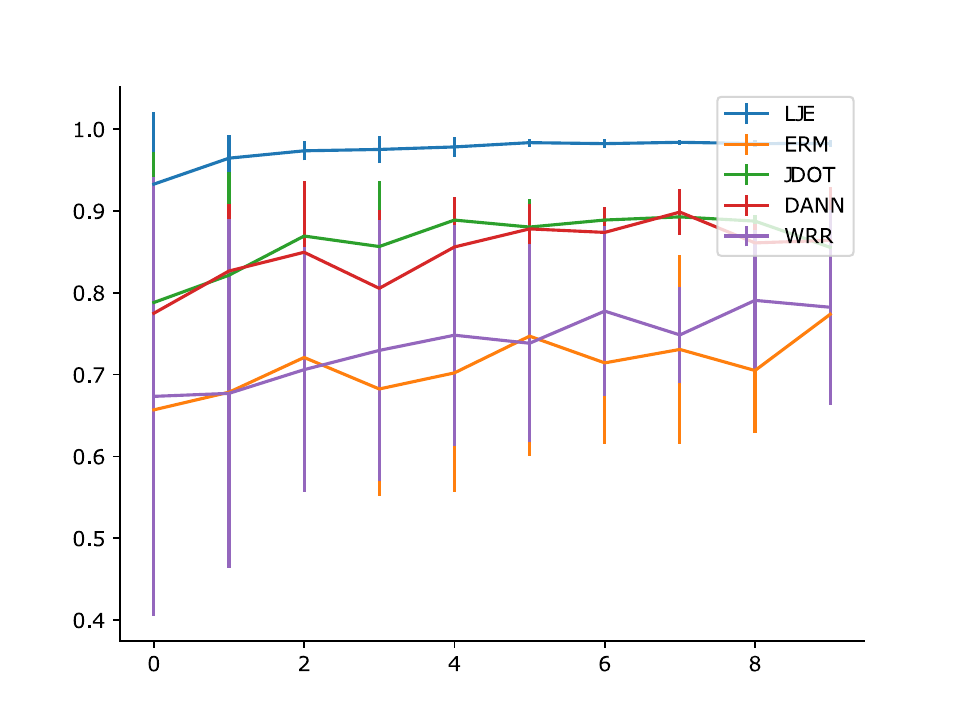}
    \end{minipage}
    \begin{minipage}{0.4\textwidth}
        \centering
        \includegraphics[width=\textwidth]{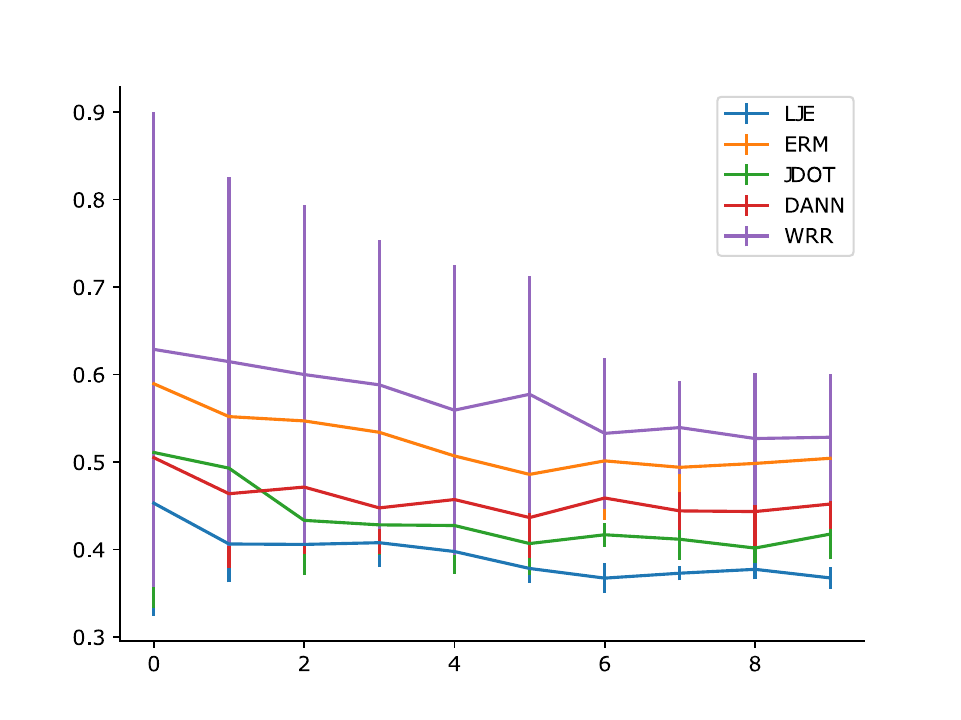}
    \end{minipage}
    \caption{Target accuracy plots for the USPS $\to$ MNIST scenario shown in Table \ref{table1} using the model conv1. Corresponding entanglement estimates are shown on the right hand figure.}
    \label{fig:acc_plots_table1_row2}
    \vspace{1em}
    \begin{minipage}{0.4\textwidth}
        \centering
        \includegraphics[width=\textwidth]{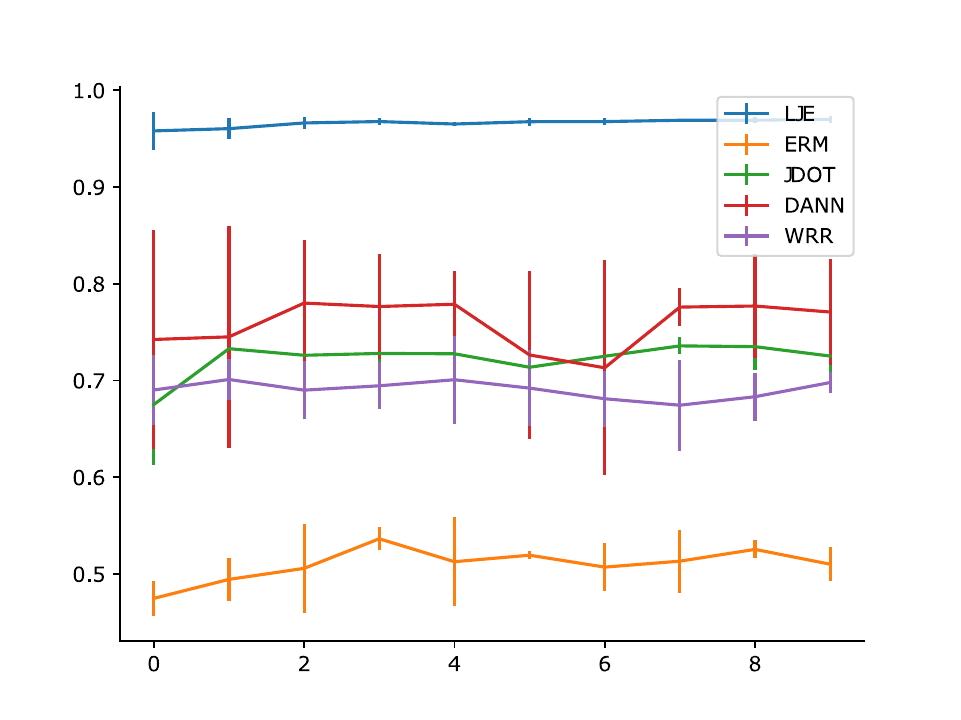}
    \end{minipage}
    \begin{minipage}{0.4\textwidth}
        \centering
        \includegraphics[width=\textwidth]{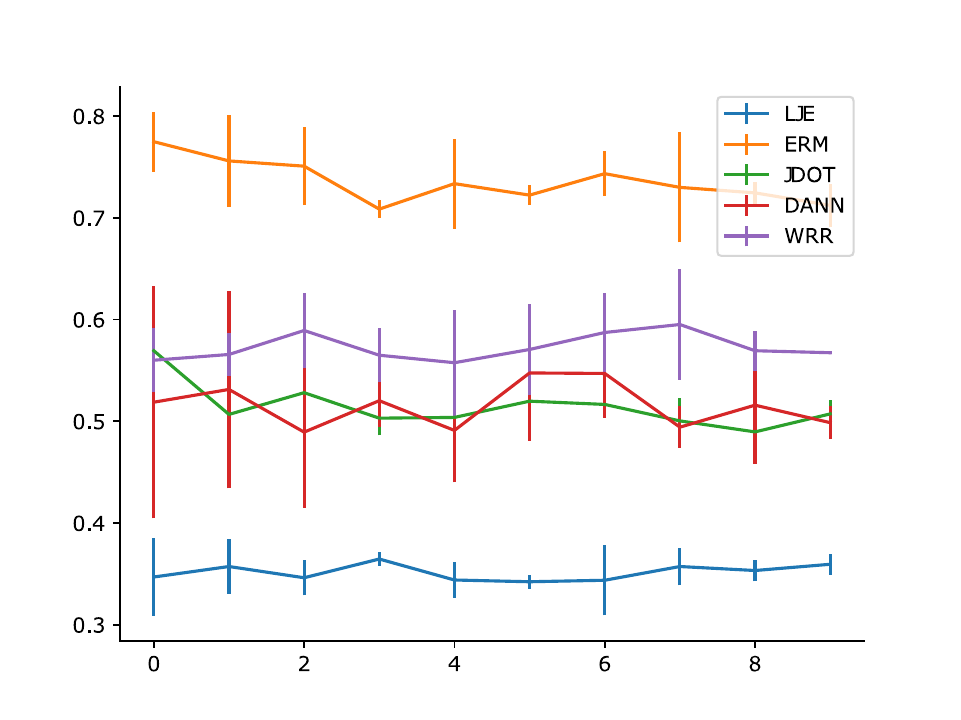}
    \end{minipage}
    \caption{Target accuracy plots for the MNIST $\to$ MNIST-M scenario shown in Table \ref{table1} using the model conv1. Corresponding entanglement estimates are shown on the right hand figure.}
    \label{fig:acc_plots_table1_row3}
\end{figure}

%% file: figures/table2/acc_plots.tex
\begin{figure}
    \centering
    \includegraphics[width=0.4\textwidth]{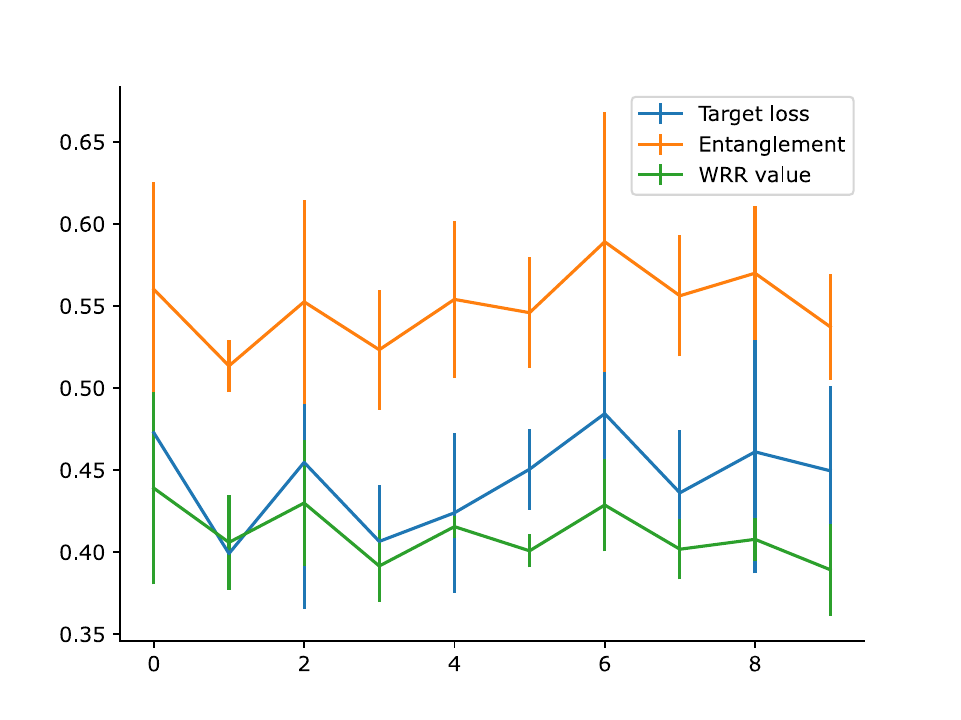}
    \caption{Target loss over the epochs for the MNIST $\to$ MNIST-M scenario shown in Table \ref{table2} using the model conv1. Corresponding entanglement estimates as well as the WRR values are included for comparison.}
    \label{fig:acc_plots_table2_row1}
    \vspace{1em}
    \centering
    \includegraphics[width=0.4\textwidth]{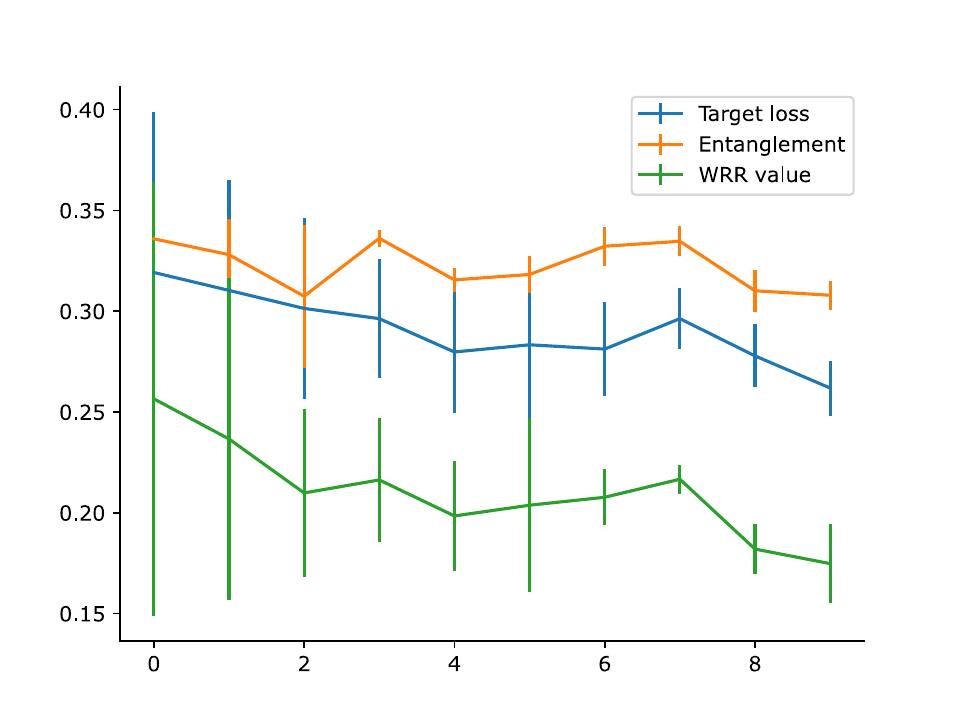}
    \caption{Target loss over the epochs for the Portraits dataset shown in Table \ref{table2} using the model MLP. Corresponding entanglement estimates as well as the WRR values are included for comparison.}
    \label{fig:acc_plots_table2_row2}
    \vspace{1em}
    \includegraphics[width=0.4\textwidth]{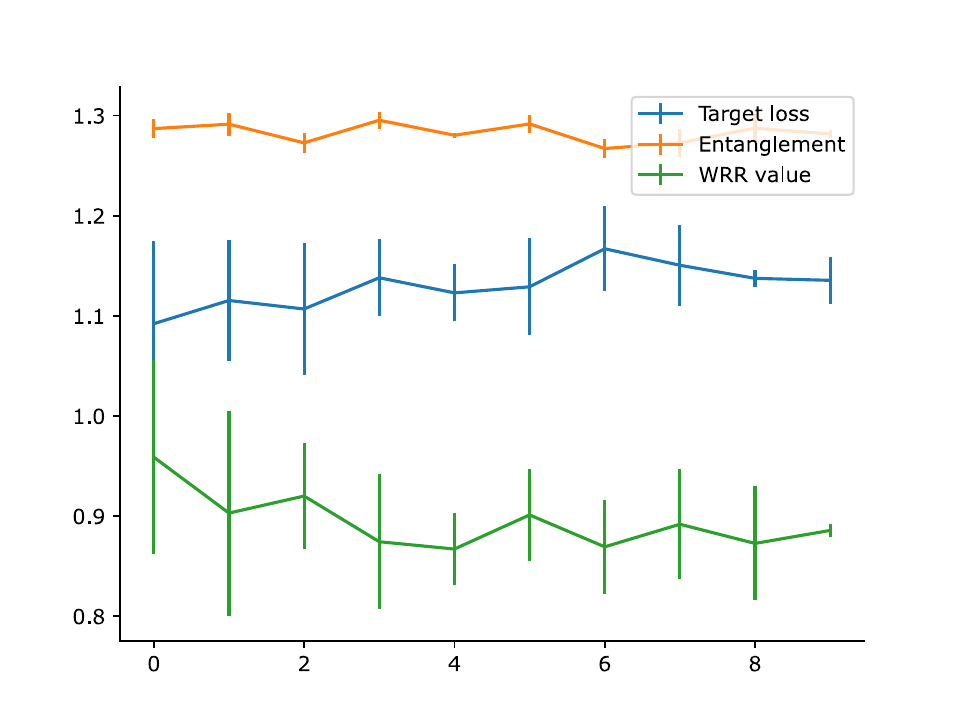}
    \caption{Target loss over the epochs for the CIFAR10 $\to$ CIFAR10c scenario shown in Table \ref{table2} using the model conv1. Corresponding entanglement estimates as well as the WRR values are included for comparison.}
    \label{fig:acc_plots_table2_row3}
\end{figure}

%% file: appendix/ablations.tex
\section{Ablations}\label{app:ablations}

In this article, we highlighted the (model, dataset, optimizer) dependency of entanglement. This triple dependency has severe consequences for UDA: changes in the target accuracy can be attributed to hyperparameters, model selection, other algorithmic choices or any preprocessing during the dataset creation (e.g., centering, label shift removal). In this section, we select several interesting ablations that highlight this manifold dependency.

\paragraph{Batch size.} We explore the change in the performance of the UDA methods as the batch size increases from $64$ to $512$ in Figures \ref{fig:ablation_batch_size_1} and \ref{fig:ablation_batch_size_2}. In the USPS to MNIST scenario, with a batch size of $64$, WRR shows a high-variance but higher accuracy performance, whereas increasing the batch size to $512$ makes the WRR performance decrease significantly. We attribute this decrease to the fact the scaling is set as $1$. JDOT with the scaling set to $0.001$ does significantly better.

\begin{figure}[h]
    \centering
    \includegraphics[width=0.3\linewidth]{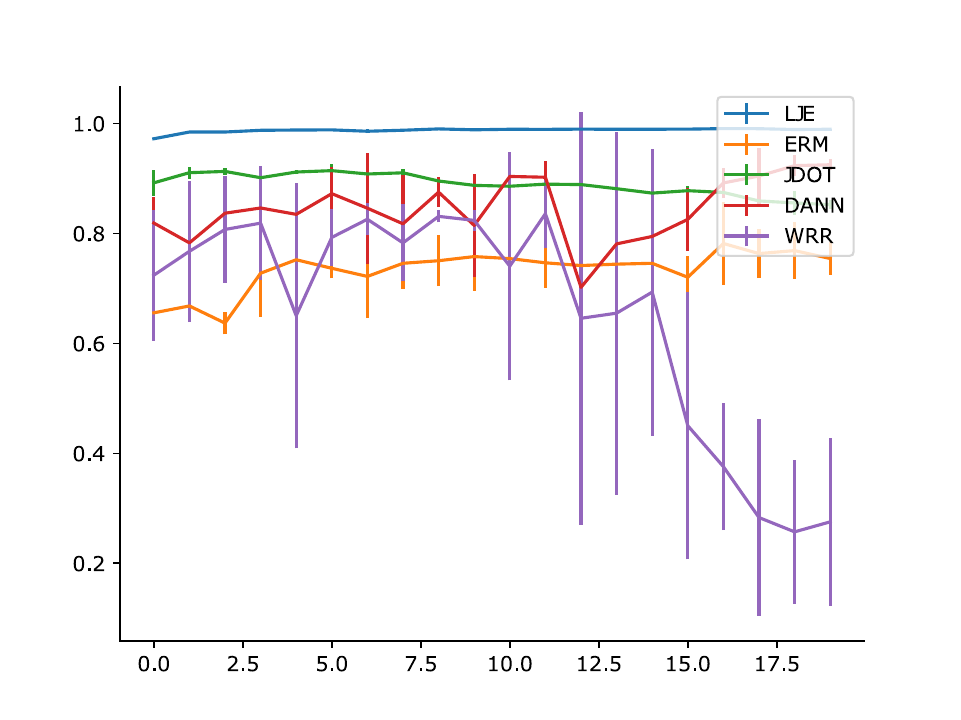}
    \caption{A case in the USPS $\to$ MNIST scenario for the conv1 model where increasing the number of epochs worsens the accuracy of the methods. All hyperparameter settings and model configurations are the same as in Table \ref{table1} except for the number of epochs.}
    \label{fig:more_epochs}
\end{figure}

\paragraph{Training for more epochs.} We find that training for more epochs can sometimes change the ranking of the methods in terms of accuracy. In Figure \ref{fig:more_epochs} showing the USPS $\to$ MNIST scenario, we see that the performance of WRR worsens, potentially due to the high variance discussed above. 

\paragraph{Effect of the loss function.} As noted in the discussion for Table \ref{table2}, to be able to relate the entanglement estimates and the WRR values during the UDA optimization to our upper bound, we switched the loss function from cross-entropy to the Euclidean distance. We explore here in more detail the change in target accuracy resulting from this switch. Table \ref{tab:ce_to_euclidean_loss} indicates that in several cases there can be significant gains in accuracy. We present the results at the fifth epoch.

\paragraph{Effect of label shift.} As noted in section~\ref{sec:experiments}, label shift deteriorates performance (potentially by increasing label-entanglement). We explore here the effect further by artificially removing label shift from two datasets. We implement class balanced sampling in the MNIST $\to$ USPS and SVHN $\to$ MNIST cases below in Table \ref{tab:label_shift} using a weighted sampler in PyTorch. We present the results at the fifth epoch. It can be seen that the entanglement estimates decrease significantly and as a result the target accuracies are significantly higher. The effect is more pronounced for the SVHN $\to$ MNIST case, where there is a label shift (in Wasserstein-1 distance) of $0.2$, as opposed to $0.13$ in the MNIST $\to$ USPS case.



\paragraph{$W_1$ vs $W_2$ distance minimization.} We used $W_2$ distance minimization in all the experiments, as we suspected that the WRR objective function may not be easy to differentiate for $\alpha \neq 2$, e.g., when $\alpha = 1$. We investigate this claim numerically by switching between the $W_2$ and $W_1$ minimization of the WRR objective function in Table \ref{tab:w1_to_w2}. WRR uses the Euclidean distance as the loss function and the accuracies are evaluated at the fifth epoch of training. From the results it is not possible to claim that one is strictly better than the other: although there is some significant gain in accuracy in the Portraits dataset and in the MNIST $\to$ USPS scenario of using $\alpha = 2$, in the USPS $\to$ MNIST scenario, we fail to see a significant gain.

\paragraph{Additional methods.} Besides the methods shown in Table 1, we also evaluated two more approaches: one based on \emph{MMD} (see, e.g., \citet{li2018domain}) and a more recent approach called \emph{Reverse-KL} \citep{nguyen2022kl}. We report in Table \ref{tab:two_more_baselines} the mean source/target accuracies obtained with the best model (conv1).

We included the ERM performance in Table \ref{tab:two_more_baselines} as well, in order to reflect on the cases of positive and negative transfer. Note that MMD does not show any cases of positive transfer. Our implementation of MMD was based on the DomainBed \citep{gulrajani2020search} repository. We attribute the uniformly poor performance to either (1) bad specification of kernel parameters or (2) our choice of the feature space. 

As for the Reverse-KL algorithm (our implementation\footnote{Note that Reverse-KL \citep{nguyen2022kl} implementation uses \emph{probabilistic representation networks}, hence our models in Table 1 had to be adapted in order to run the algorithm.} is a modification of the code in the Reverse-KL repository), note the uneven performance: Reverse-KL replaces DANN and JDOT methods as the best method in the scenarios MNIST $\to$ USPS, USPS $\to$ MNIST and Portraits, yet drops down to chance levels in others. We attribute this to our choice of hyperparameters ($\beta = 0.1$ and $\beta_{\textrm{aux}} = 0.1$ in \citet{nguyen2022kl}): by varying them we can improve in the scenarios with poor results but then the accuracies in the above mentioned scenarios get significantly worse. Note that, similarly for JDOT \citep{damodaran2018deepjdot}, we observe much better results by changing hyperparameters between experiments. This excessive hyperparameter-dependency is an instance of the (model, optimizer, dataset) triplet dependency of the entanglement that we tried to highlight in this article.

\paragraph{Additional dataset and model.} We looked at an additional dataset called \emph{Office-Home} \citep{venkateswara2017deep} which consists of four different subdatasets (Art, Clipart, Product and Real World) with corresponding domain adaptation scenarios (e.g., Art $\to$ Clipart). The models that we tried in Table 1 are unable to get high accuracy in these challenging datasets (e.g., LJE and CCA accuracies are very low) due to the high resolution of the images, small dataset size and wide distribution shifts observed between the subdatasets, among other potential difficulties. As a remedy, we used an ImageNet-pretrained \emph{ResNet18} \cite{he2016deep} as our model (with the last layer suitably changed to output $65$ classes). We used the same settings as reported in the experiments, except for lowering the Adam learning rate to $1 \times 10^{-4}$. We provide in Table \ref{tab:office_home} some preliminary results on the Product $\to$ Real World scenario (denoted as Pr $\to$ Rw).

From our preliminary results, it seems that JDOT is the best performing algorithm here (and the only one that shows positive transfer, comparing to ERM target accuracy). Interestingly WRR, which can be seen as a bare-bones version of JDOT (without the intermediate feature space alignment and without the scaling) performs poorly here, which highlights again the dependence on hyperparameter selection, as part of the optimizer component of entanglement's (model, optimizer, dataset) triplet dependency.

\clearpage

\begin{table}[b!]
    \centering
    \caption{The effect of switching from cross entropy to the Euclidean distance as the WRR loss function.}
    \vspace{0.1cm}
    \begin{tabular}{c|c|c}
        \toprule
        Scenario / Model & Cross-entropy Acc. & Euclidean distance Acc.\\
        \midrule
        MNIST $\to$ USPS / small$\_$cnn & 86.8 $\pm$ 1.0 & 94.7 $\pm$ 1.1 \\
        USPS $\to$ MNIST / LeNet & 72.6 $\pm$ 17.3 & 93.5 $\pm$ 2.0 \\
        SVHN $\to$ MNIST / small$\_$cnn & 41.2 $\pm$ 0.4 & 57.3 $\pm$ 10.7 \\
        MNIST $\to$ MNIST-M / MLP & 12.2 $\pm$ 1.1 & 16.7 $\pm$ 1.4 \\
        \bottomrule
    \end{tabular}
    \label{tab:ce_to_euclidean_loss}
\end{table}

\begin{table}[h]
    \centering
    \caption{The effect of class-balanced sampling during the WRR optimization.}
    \vspace{0.1cm}
    \begin{tabular}{c|c|c|c}
        \toprule
        Scenario / Model & $W_{1}(\hat{p}_y, \hat{q}_y)$ & Unbalanced Acc. / Entanglement & Balanced Acc. / Entanglement \\
        \midrule
        MNIST $\to$ USPS / LeNet & 0.13 / 0.0 & 84.7 $\pm$ 2.7 / 0.45 $\pm$ 0.02 & 91.4 $\pm$ 0.7 / 0.35 $\pm$ 0.01 \\
        SVHN $\to$ MNIST / conv1 & 0.2 / 0.0 & 83.3 $\pm$ 7.7 / 0.51 $\pm$ 0.06 & 95.3 $\pm$ 0.83 / 0.38 $\pm$ 0.01 \\
        \bottomrule
    \end{tabular}
    \label{tab:label_shift}
\end{table}

\begin{table}
    \centering
    \caption{The effect of switching from $W_2$ to $W_1$.}
    \vspace{0.1cm}
    \begin{tabular}{c|c|c}
        \toprule
        Scenario / Model & $W_{2}$-minimization Acc. & $W_{1}$-minimization Acc. \\
        \midrule
        MNIST $\to$ USPS / small$\_$cnn & 94.7 $\pm$ 1.1 & 92.3 $\pm$ 0.29 \\
        USPS $\to$ MNIST / LeNet & 95.3 $\pm$ 0.4 & 93.5 $\pm$ 2.0 \\
        Portraits 1905--1974 $\to$ 1974--2013 / conv2 & 84.07 $\pm$ 1.05 & 80.0 $\pm$ 1.0 \\
        \bottomrule
    \end{tabular}
    \label{tab:w1_to_w2}
\end{table}

\begin{table}
    \centering
    \caption{The source/target accuracy of two other UDA methods (with ERM performance added as a baseline).}
    \vspace{0.1cm}
    \begin{tabular}{c|c|c|c}
        \toprule
        Scenario / Model & ERM & MMD &  Reverse-KL \\
        \midrule
        MNIST $\to$ USPS / conv1 & 98.0 / 83.8 & 98.5 / 80.7 & 98.9 / 90.2 \\
        USPS $\to$ MNIST / conv1 & 98.9 / 83.8 & 95.7 / 48.2 & 96.4 / 97.3 \\
        MNIST $\to$ MNIST-M / conv1 & 98.9 / 47.5 & 98.7 / 44.0 & 97.7 / 10.3 \\
        SVHN $\to$ MNIST / conv1 & 91.2 / 64.9 & 89.0 / 61.3 & 19.4 / 11.2 \\
        Portraits / conv1 & 96.8 / 86.3 & 95.5 / 79.1 & 96.7 / 87.7 \\
        CIFAR10 $\to$ CIFAR10c / conv1 & 75.5 / 37.7 & 62.6 / 30.9 & 21.1 / 10.7 \\
        \bottomrule
    \end{tabular}
    \label{tab:two_more_baselines}
\end{table}

\begin{table}
    \centering
    \caption{The source/target accuracy of different methods using an ImageNet-pretrained ResNet18 model on the Office-Home dataset, Product $\to$ Real World distribution shift scenario.}
    \vspace{0.1cm}
    \begin{tabular}{c|c|c|c|c|c|c}
        \toprule
        Scenario / Model & LJE & ERM & WRR & JDOT & DANN & MMD \\
        \midrule
        Office-Home Pr $\to$ Rw & 84.4 / 76.9 & 86.5 / 54.0 & 2.8 / 1.7 & 88.5 / 63.9 & 84.4 / 48.0 & 84.1 / 53.7 \\
        \bottomrule
    \end{tabular}
    \label{tab:office_home}
\end{table}

\clearpage

\input{figures/ablations/batch_size}

%% file: figures/ablations/batch_size.tex
\begin{figure}[p]
    \centering
    \begin{minipage}{0.3\textwidth}
        \centering
        \includegraphics[width=\textwidth]{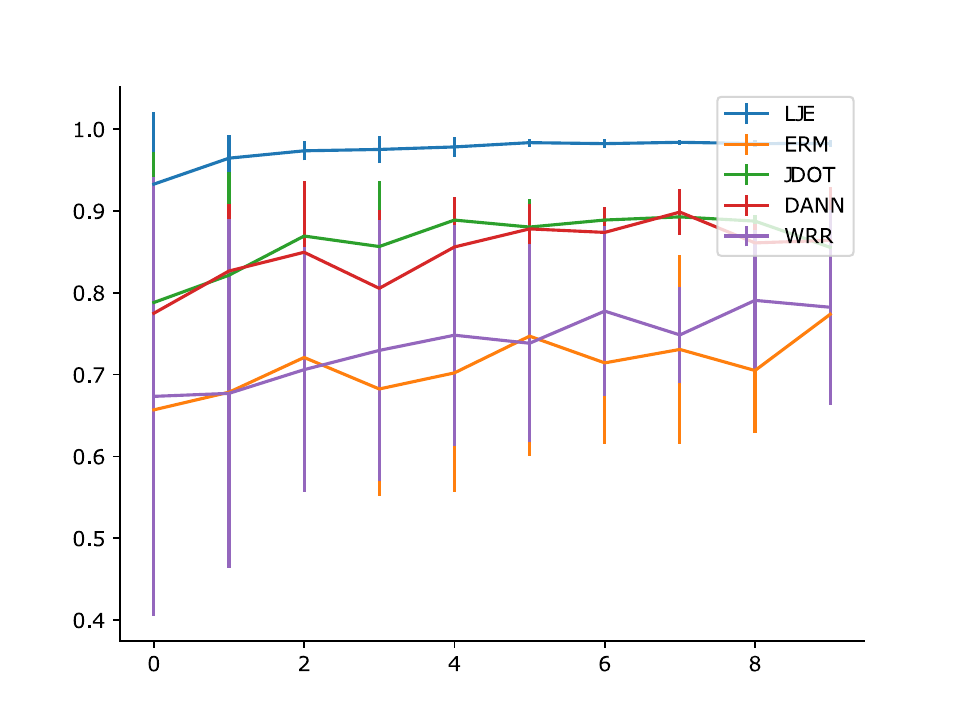}
    \end{minipage}
    \begin{minipage}{0.3\textwidth}
        \centering
        \includegraphics[width=\textwidth]{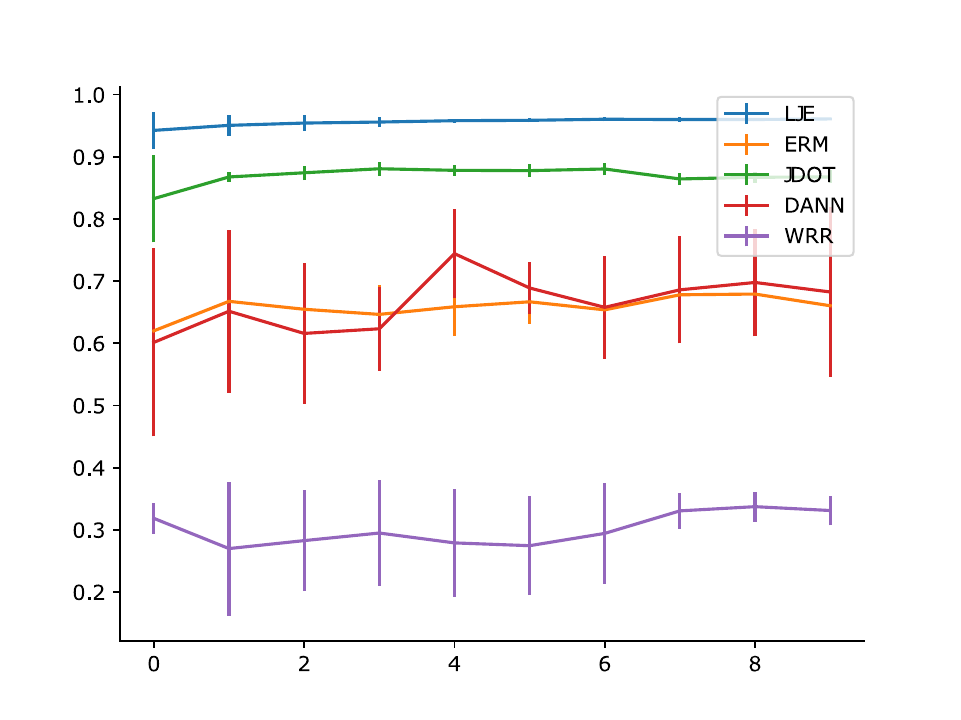}
    \end{minipage}
    \caption{Target accuracy plots for the USPS $\to$ MNIST scenario shown in Table \ref{table1} using the model conv1. Left/right hand sides correspond to batch sizes $64$ and $512$ respectively.}
    \label{fig:ablation_batch_size_1}
    \vspace{1em}
    \centering
    \begin{minipage}{0.3\textwidth}
        \centering
        \includegraphics[width=\textwidth]{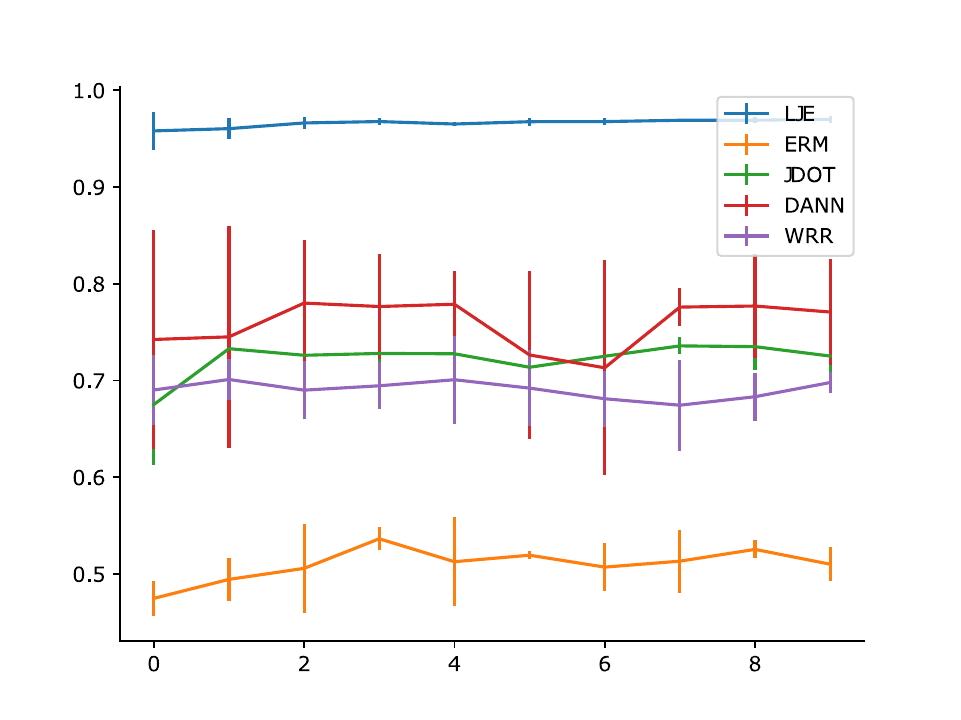}
    \end{minipage}
    \begin{minipage}{0.3\textwidth}
        \centering
        \includegraphics[width=\textwidth]{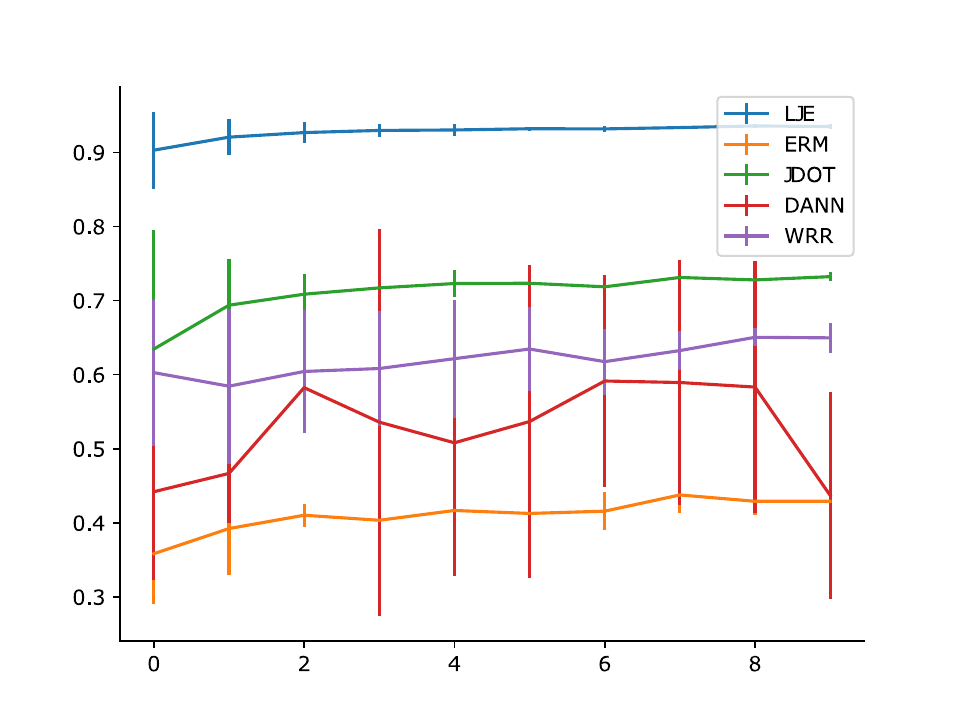}
    \end{minipage}
    \caption{Target accuracy plots for the MNIST $\to$ MNIST-M scenario shown in Table \ref{table1} using the model conv1. Left/right hand sides correspond to batch sizes $64$ and $512$ respectively.}
    \label{fig:ablation_batch_size_2}
\end{figure}